\documentclass[10pt,journal,compsoc]{IEEEtran}
\usepackage{amsmath,amssymb,amsfonts}
\usepackage{algorithmic}
\usepackage{algorithm}
\usepackage{graphicx}
\usepackage{subfigure}
\usepackage{textcomp}
\usepackage{xcolor}
\usepackage{booktabs}
\usepackage{tabularx}
\usepackage{url}

\newcommand{\Input}{\item[\textbf{Input:}]}
\newcommand{\Output}{\item[\textbf{Output:}]}

\ifCLASSOPTIONcompsoc
  \usepackage[nocompress]{cite}
\else
  \usepackage{cite}
\fi
\ifCLASSINFOpdf
\else
\fi
\begin{document}
%
\title{Pareto-Optimal Estimation and Policy Learning on Short-term and Long-term Treatment Effects}
%
%
%
%

\author{Yingrong Wang, Anpeng Wu, Haoxuan Li~\IEEEmembership{Member,~IEEE,}
Weiming Liu, \\ Qiaowei Miao, Ruoxuan Xiong, Fei Wu~\IEEEmembership{Senior Member,~IEEE,}
Kun Kuang
\IEEEcompsocitemizethanks{\IEEEcompsocthanksitem Yingrong Wang, Anpeng Wu, Weiming Liu, Qiaowei Miao, Fei wu, Kun Kuang are with the College of Computer Science and Technology, Zhejiang University, Hangzhou, Zhejiang Province 310027, China. E-mail: \{wangyingrong, anpwu, 21831010, QiaoweiMiao, kunkuang\}@zju.edu.cn and wufei@cs.zju.edu.cn.

\IEEEcompsocthanksitem Haoxuan Li is with the Center for Data Science, Peking University, Beijing, China. E-mail: hxli@stu.pku.edu.cn.

\IEEEcompsocthanksitem Ruoxuan Xiong is with the Department of Quantitative Theory \& Methods, Emory University, USA. E-mail: ruoxuan.xiong@emory.edu.

\IEEEcompsocthanksitem Correspondence to: Kun Kuang. Yingrong Wang and Anpeng Wu contributed equally to this work.
}

\thanks{Manuscript received April 19, 2005; revised August 26, 2015.}}

%
%

\markboth{Journal of \LaTeX\ Class Files,~Vol.~14, No.~8, August~2015}%
{Shell \MakeLowercase{\textit{et al.}}: Bare Demo of IEEEtran.cls for Computer Society Journals}
%



\IEEEtitleabstractindextext{%
\begin{abstract}
This paper focuses on developing Pareto-optimal estimation and policy learning to identify the most effective treatment that maximizes the total reward from both short-term and long-term effects, which might conflict with each other. For example, a higher dosage of medication might increase the speed of a patient's recovery (short-term) but could also result in severe long-term side effects. Although recent works have investigated the problems about short-term or long-term effects or the both, how to trade-off between them to achieve optimal treatment remains an open challenge. Moreover, when multiple objectives are directly estimated using conventional causal representation learning, the optimization directions among various tasks can conflict as well. In this paper, we systematically investigate these issues and introduce a Pareto-Efficient algorithm, comprising Pareto-Optimal Estimation (POE) and Pareto-Optimal Policy Learning (POPL), to tackle them. POE incorporates a continuous Pareto module with representation balancing, enhancing estimation efficiency across multiple tasks. As for POPL, it involves deriving short-term and long-term outcomes linked with various treatment levels, facilitating an exploration of the Pareto frontier emanating from these outcomes. Results on both the synthetic and real-world datasets demonstrate the superiority of our method.
\end{abstract}

}

\maketitle

\IEEEdisplaynontitleabstractindextext

%
\IEEEpeerreviewmaketitle

\section{Introduction}
\begin{figure*}[t]
\centering
\includegraphics[width=1\textwidth]{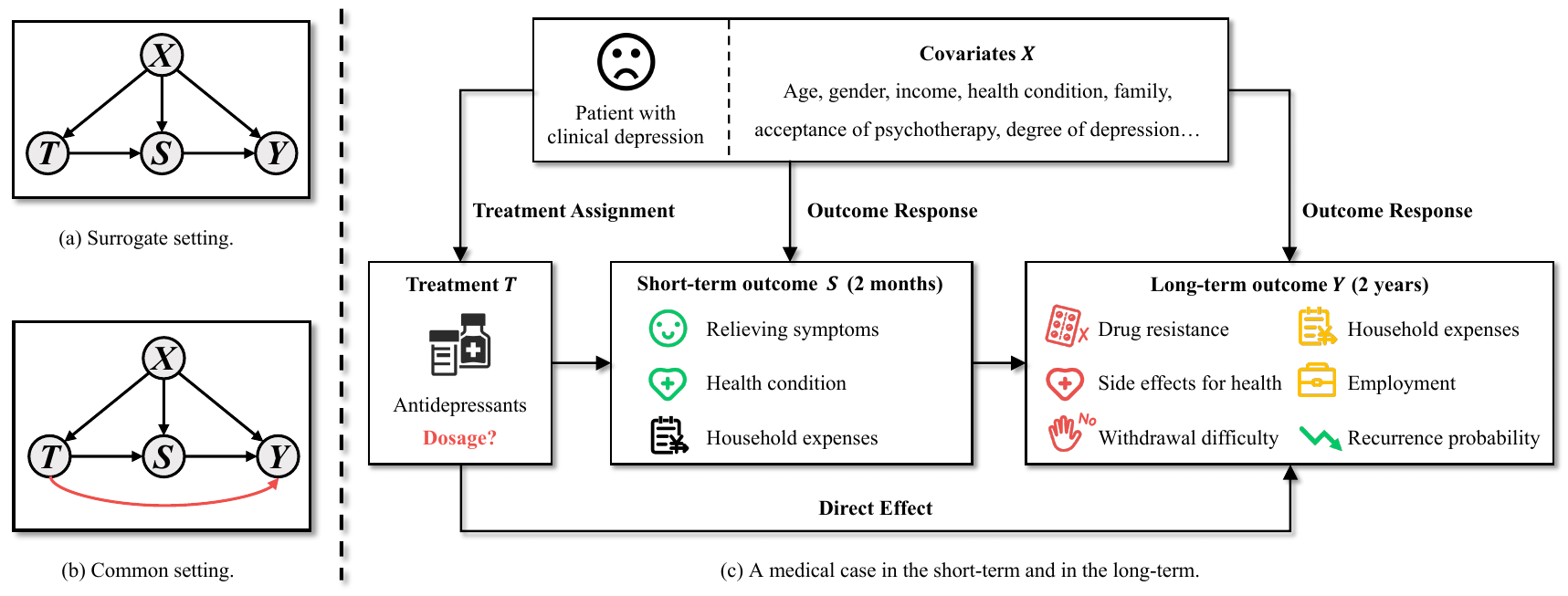}
\caption{Illustration of settings in the long-term treatment effect estimation. (a) In the surrogate setting, the short-term outcome $S$ serves as a mediator to block the effect of treatment $T$ on the long-term outcome $Y$. (b) In the common setting, the direct influence from $T$ to $Y$ is considered and highlighted in red. (c) We give a medical case to illustrate the importance of trade-off between the short-term and long-term outcomes that conflict with each other.}
\label{fig:causal-graph}
\end{figure*}

In causal inference and policy learning, the estimation of the causal effects, both in the short and long term, is a crucial concern across various fields such as healthcare, education, marketing, and social science~\cite{hu2023identification,chetty2011education,yang2020targeting}.
For example, when considering the dosage of antidepressants for depression, as illustrated in Fig.~\ref{fig:causal-graph}, typically, researchers and practitioners focus on short-term indicators such as symptom relief, health condition improvements, and household expenses within the first two months. These short-term effects are more manageable to study as they appear within days or months. However, long-term outcomes are also crucial. These include drug resistance and side effects that can emerge after two years, potentially affecting the patient's life and employment prospects. Unfortunately, these long-term effects are rarely observed and studied due to the high costs and extended time frames required for long-term studies. This gap in research highlights a crucial area of studying short-term and long-term causal effects in policy learning and causal inference. 

Recently, researchers have developed methods to estimate both the short-term and long-term outcomes under the potential outcome framework~\cite{rubin1974estimating}. Controlling confounding bias in this scenarios has been extensively discussed, involving various approaches like propensity-based methods~\cite{stratification,re-weighting,matching}, balancing methods~\cite{entropy,approximate}, representation-based methods~\cite{BLR,CFR}, generative modeling methods~\cite{GANITE,CEVAE}, etc. Athey et al. \cite{athey2019surrogate} initiated the exploration of long-term outcome estimation under the surrogate framework, as depicted in Fig.~\ref{fig:causal-graph}(a). This framework posits that the long-term outcome is independent of the treatment, given the short-term outcome. Consequently, the short-term outcome is conceptualized as a surrogate or mediator for the long-term outcome. This innovative approach has inspired subsequent research, as seen in various studies~\cite{athey2020combining,chen2023semiparametric,kallus2022role}, which further elaborate and expand upon the surrogate framework~\cite{freedman1992statistical,frangakis2002principal,joffe2009related}. However, these works overlook the direct effect of treatment on long-term outcomes. This aspect is crucial and more commonly encountered in real-world scenarios, as illustrated in Fig.~\ref{fig:causal-graph}(b,c).
While researchers~\cite{hu2023identification,imbens2023longterm,ghassami2022combining} have focused on eliminating confounding bias and estimating potential outcomes, these methods rely on a data fusion containing the random trial data from control experiments. Another innovative perspective \cite{norcliffe2023survivalgan} regards the long-term outcomes as latent variables and recovers them using the short-term outcomes by generative models~\cite{GAN}.

Although many works have investigated the treatment effect estimation on short-term and long-term outcomes, the problem of conflicts between them is rarely discussed. A pertinent example is the administration of medication: a higher dose may accelerate short-term recovery yet potentially cause serious long-term side effects. Moreover, the conflicts between short- and long-term outcomes have another meaning, i.e. the optimization directions for various tasks can also conflict when training estimation models. Therefore, how to trade-off between the short-term and long-term causal effects to achieve optimal treatment remains an open challenge. Last but not least, previous works mainly focus on binary treatment cases and directly regressing outcomes together without fully utilizing the information~\cite{kohavi2012trustworthy}. However, the nuanced scenario of continuous treatments and their impact on multiple outcomes is less explored. This gap is notable in the context of confounder balancing and policy learning. When dealing with continuous treatments, for instance, the application of balanced representation learning to estimate multiple outcomes concurrently can lead to significant information loss.

In this study, we comprehensively investigate the complexities of optimizing treatments and their impacts on multiple outcomes, introducing a novel Pareto-Efficient algorithm that consists of two components: Pareto-Optimal Estimation (POE) and Pareto-Optimal Policy Learning (POPL). POE integrates a continuous Pareto module along with representation balancing, which significantly improves the efficiency of estimation across various tasks. On the other hand, POPL focuses on deriving both short-term and long-term outcomes associated with different levels of treatment. This approach aids in exploring the Pareto frontier that emerges from these outcomes, providing a comprehensive understanding of the impacts and trade-offs involved in treatment optimization.

The contributions are three-folds.
\begin{itemize}
    \item We address the new challenge of estimation and policy learning tasks on the short-term and long-term treatment effects with conflicts, which is beyond the capability of previous methods.
    \item We propose a novel Pareto-efficient algorithm that contains two modules named POE and POPL. It can offer not only accurate estimations of short-term and long-term outcomes but also maximal reward from effective policy learning.
    \item We validate our method through extensive experiments, demonstrating its superiority across five diverse datasets, including one real-world dataset, one simulated dataset, and three semi-synthetic datasets.
\end{itemize}

\section{Related Work}

Causal inference is a powerful tool in data-related fields, enabling a deeper understanding and explanation of the complex relationships between data, such as in recommendation systems~\cite{lu2021long-term,zhu2023dcmt}, network embedding~\cite{network-embedding}, and SQL queries~\cite{chu2023continual, wang2023sequential}. It assists in revealing the causal dynamics behind query results, offering more profound insights into the data~\cite{shen2023causal, youngmann2023explaining}.
Furthermore, the significance of multi-term outcome responses and policy learning in the data management community cannot be overstated \cite{multi-task-survey, davoudi2021policy, zhao2023personalized, chen2023rec}. These topics reflect the rapidly evolving landscape of data science and analytics, highlighting the field's continuous advancement. Next, we will further discuss three directions that are highly relevant to this paper.

\subsection{Long-term Treatment Effect Estimation}
In the field of causal inference, Susan et al. pioneered the exploration of long-term treatment effect estimation, as documented in their seminal work~\cite{athey2019surrogate}. Their approach utilized short-term outcomes as mediators, assuming independence of long-term outcomes from the treatment given these mediators. This paradigm has been further developed in subsequent studies~\cite{athey2020combining,chen2023semiparametric,kallus2022role}. The core objective of these methodologies is to harmonize data from diverse datasets using a variety of balancing scores, akin to the propensity score method established by Rosenbaum and Rubin~\cite{rosenbaum1983central}. These balancing scores serve as crucial tools for aligning datasets in a manner that facilitates more accurate and reliable estimation of treatment effects.
However, the challenge of selecting valid short-term surrogates, that fully mediate the effect of the treatment, has been a subject of extensive debate over the years~\cite{prentice1989surrogate,frangakis2002principal,lauritzen2004discussion,chen2007criteria,ju2010criteria}. Additionally, there exists a phenomenon known as the surrogate paradox~\cite{chen2007criteria}, where the treatment’s effect on both the short-term surrogate and the long-term outcome is positive, yet the treatment adversely affects the outcome of interest. Consequently, many researchers have shifted focus to a broader context as illustrated in Fig.~\ref{fig:causal-graph}, considering the direct effect of treatment on the long-term outcome.

Additionally, there is a growing body of research focused on causal inference using long-term data. A notable example is the Long-Term Effect Estimation (LTEE) approach proposed by Lu et al. in \cite{lu2021long-term}, which is designed to learn surrogate representations for estimating causal effects with sequential outcomes. Diverging from the traditional setting of long-term effect estimation, Chu et al. in \cite{chu2023continual} propose a novel perspective where observational data is considered incremental, reflecting the dynamic changes over time. Moreover, it's posited that sequential outcome data can be interpreted as an alternative form of long-term outcome, broadening the scope of analysis in causal inference studies.

\subsection{Multi-task Learning}
The short-term and long-term treatment effect estimation also could be regarded as multi-task learning. Sener et al. introduced the Multi-Gradient Descent Algorithm (MGDA) for modeling multi-task learning as a multi-objective optimization problem \cite{sener2018multi}. This algorithm ensures that solutions are either on the Pareto boundary or represent optimal directions for simultaneous task improvement. A notable limitation of MGDA, however, is its provision of a singular solution point, which may not meet the varied demands of practical applications. To address this, the concept of Pareto MTL was introduced by Lin et al. \cite{lin2019pareto}, which deconstructs the multi-objective optimization challenge into a series of constrained sub-problems, each epitomizing different trade-off preferences. This approach allows for the derivation of a diverse set of Pareto optimal solutions. Nonetheless, Pareto MTL requires individual training for each solution, thus not fully exploiting the continuous nature of the Pareto frontier. Addressing this shortfall, Ma et al. proposed an efficient methodology in \cite{ma2020efficient} that starts from an initial Pareto solution and progressively identifies additional solutions, leveraging the continuity of the Pareto frontier. This approach, while comprehensive, incurs significant computational complexity. To enhance computational efficiency, the XWC-MGDA algorithm was developed \cite{momma2022multi}, facilitating exploration of the Pareto frontier from any chosen reference point. Additionally, a comprehensive survey on multi-task learning \cite{multi-task-survey} presents theoretical insights and outlines several prospective avenues for future research in this domain.

\subsection{Policy Learning}
In this work, we still focus on policy learning for Pareto-optimal policy to trade-off between the short-term and long-term outcomes, with the purpose of maximizing the total reward. Numerous studies have been dedicated to policy learning from observational data, which can be broadly categorized into two main streams: statistics~\cite{luedtke2016statistical,qian2011performance,zhang2012estimating} and machine learning~\cite{beygelzimer2009offset,swaminathan2015batch,kallus2018balanced}. The first category primarily tackles empirical maximization problems and delves into their various relaxations, aiming to refine the theoretical underpinnings of policy learning. The second category, on the other hand, concentrates on enhancing the practical performance of policy learning techniques, with a particular emphasis on the application of doubly robust objectives. This bifurcation of focus not only delineates the diverse methods employed in policy learning but also underscores the multifaceted nature of this area.
In this area, the study of adaptive paywalls in \cite{davoudi2021policy} explores the balance between user satisfaction and cost. Similarly, research in \cite{zhao2023personalized} presents a strategy to reduce revenue loss from ad blockers. Additionally, the Sim2Rec framework \cite{chen2023rec}, used in sequential recommendation systems, focuses on policies that increase user engagement for long-term benefits. Together, these studies emphasize the importance of balancing different outcomes in policy learning.

\section{Problem Setup}
\label{sec:problem}

\subsection{Notations and Assumptions}
In this paper, we focus on estimating both the long-term and short-term treatment effects to provide a thorough comprehension of the treatment's total effect on desired outcomes from observations, and then to find an optimal treatment in trade-off between these two results. In the observational data $\mathbb{D}=\{X_i,T_i,S_i,Y_i\}_{i=1}^n$, for each unit $i$ with covariates $X_i \in \mathbb{X}$ where $\mathbb{X} \subset  \mathbb{R}^{m_X}$, we observe a continuous treatment variable $T_i \in \mathbb{T}$ where $\mathbb{T}\subset  \mathbb{R}$ and two outcome variables $S_i \in \mathbb{Y}$ for short-term outcome and $Y_i \in \mathbb{Y}$ for long-term outcome where $\mathbb{Y} \subset \mathbb{R}$. As depicted in Fig.~\ref{fig:causal-graph}, in the studies of short- and long-term causal effects, the individual preference and attributes, i.e. $X_i$, would decide the treatment choice $T_i$ and affect the two outcomes $S_i$ and $Y_i$ simultaneously. Then, the treatment $T_i$ would also have direct effects on $S_i$ and $Y_i$. Under the potential outcome framework~\cite{rubin1974estimating}, we expect to accurately estimate the short-term outcome $S(t)$ while also hoping to enhance the accuracy of our estimates for the long-term outcome $Y(t)$ for any assigned treatment $t$:
\begin{equation}
    \mathbb{E}[S(t)|X=x]=\mathbb{E}[S|X=x,T=t]
\end{equation}
\begin{equation}
    \mathbb{E}[Y(t)|X=x,S=s]=\mathbb{E}[Y|X=x,T=t,S=\hat{s}].
\end{equation}

To identify the above potential outcomes for any treatments, we also require the following assumptions: 
\begin{enumerate}
    \item \textit{Consistency.} If an individual receives a treatment $ t $ from set $\mathcal{T}$, their observed outcomes $ s $ and $ y $ are identical to the potential outcomes $ S(t) $ and $ Y(t) $, i.e. $S(t)=s$ and $Y(t)=y$ when the assigned treatment is $T=t$. 
    \item \textit{Unconfoundedness.} The potential outcomes $ S(t) $ and $ Y(t) $ are independent of the treatment assignment $ T $, given the covariates $ X $,
    i.e. $S(t),Y(t) \perp T|X$, $\forall{{t}}\in\mathcal{T}$.
    \item \textit{Overlap.} For every treatment $ t $ in $\mathcal{T}$, there is always a positive probability of receiving that treatment given the covariates $ X $, i.e. $\mathbb{P}(T=t|X)>0, \forall t\in\mathcal{T}$.
    \item \textit{Smoothness.} The potential outcomes $ S(t) $ and $ Y(t) $ change gradually and predictably as the treatment $ T $ changes. In other words, $S(t)$ and $Y(t)$ are smooth responses to the treatment $T=t$.
\end{enumerate}
Then, in learning the balance representation, we would like to combine the information of short-term outcomes and long-term outcomes to promote the learning of the representation of $X$, as the additional information will help us learn a more valuable balanced representation. Then, we can use the estimated potential outcomes to help us find the optimal treatment for both outcomes, which would be introduced in the next subsection.

\subsection{Preliminary}

\textbf{{Pareto-Optimal Estimation (POE).} }
However, there might be conflicts in the learning of potential short-term outcomes and long-term outcomes, leading to either the information of long-term results dominating or the information of short-term results prevailing, causing the representation to not only fail in integrating the learning information of both outcomes but also to be biased by the information of another outcome, resulting in an overall decline in model performance.
We can reformulate them into multi-task Pareto regression, including confounding balancing constraints, short-term outcome regression, long-term outcome regression, and so on.
In the objective function $\mathcal{L}(\boldsymbol{\theta})=[\mathcal{L}_1(\boldsymbol{\theta}),\dots,\mathcal{L}_m(\boldsymbol{\theta})]^T$ with $m$ tasks (specific definitions in our model can be referred to as Eq. (\ref{eq:mi}), Eq. (\ref{eq:mse-s}), Eq. (\ref{eq:regret_s}), and Eq. (\ref{eq:regret_y})), $\boldsymbol{\theta}\in\mathbb{R}^n$ represents the parameters of a backbone to fulfill these tasks.

Thus, our objective is simply to find optimal representation networks with parameters $\boldsymbol{\theta}^*$ so as to integrate all available information to enhance the performance, i.e. $\min\limits_{\boldsymbol{\theta}^*\in\mathbb{R}^n}\mathcal{L}(\boldsymbol{\theta}^*)$. However, it is hard to achieve such $\boldsymbol{\theta}^*$ due to the conflicts of the optimization directions among multiple tasks, and thus we transfer it as Pareto estimation problem~\cite{zitzler1999multi,lv2021paretoDA,momma2022multi}: 
\begin{enumerate}
    \item \textit{Pareto dominance.} We say that a solution $\boldsymbol{\theta}'$ dominates another solution $\boldsymbol{\theta}$ if $\mathcal{L}_i(\boldsymbol{\theta}')\le\mathcal{L}_j(\boldsymbol{\theta})\ \forall i\in[m]$ and $\mathcal{L}_i(\boldsymbol{\theta}')<\mathcal{L}_j(\boldsymbol{\theta})\ \exists j\in[m]$. For simplicity, we denote $\boldsymbol{\theta}'$ dominating $\boldsymbol{\theta}$ as $\boldsymbol{\theta}'\vartriangleleft\boldsymbol{\theta}$, and $\boldsymbol{\theta}'\not\vartriangleleft\boldsymbol{\theta}$ otherwise.
    \item \textit{Pareto optimality.} The solution $\boldsymbol{\theta}^*$ is Pareto optimal if there is no solution dominating it. Formally, $\boldsymbol{\theta}\not\vartriangleleft\boldsymbol{\theta}^*\ \forall\boldsymbol{\theta}\in\mathbb{R}^n-\{\boldsymbol{\theta}^*\}$.
    \item \textit{Pareto frontier.} All the Pareto optimal solutions comprise of the Pareto frontier.
\end{enumerate}
Inspired by \cite{lv2021paretoDA}, we apply a continuous Pareto optimization algorithm to update the parameters of networks, and explore the optimal representation on the Pareto frontier for potential outcomes estimation. 

\textbf{{Pareto-Optimal Policy Learning (POPL).} }
In this study, our focus extends to the identification of an optimal treatment strategy that adeptly balances the trade-off between short-term and long-term outcomes. The overarching goal is to maximize the overall reward derived from the treatment. To approach this challenge, we reconceptualize the issue of finding a balance between these outcomes as a Pareto optimal problem. This reformation requires the formulated policy to seek out an optimal solution positioned along the Pareto frontier. The intention is to ensure that no other feasible solution could improve one type of outcome (short-term or long-term) without compromising the other. This process of finding and implementing such a solution is what we refer to as Pareto-Optimal Policy Learning.

\begin{figure*}[t]
\centering
\includegraphics[width=1\textwidth]{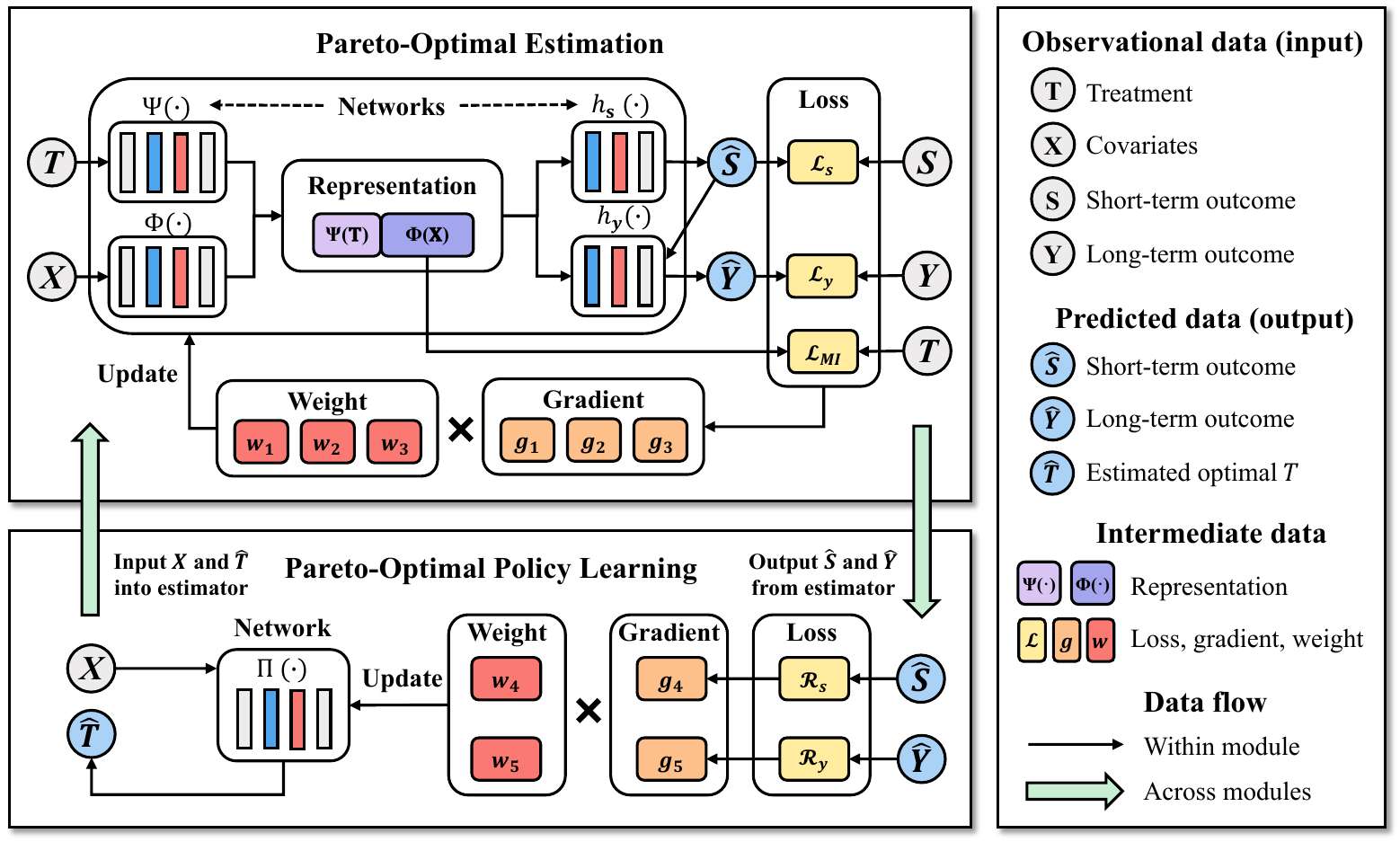}
\caption{The architecture of our model. Observational variables are marked in grey and the results predicted by our models are marked in blue. Intermediate data includes the learned representations ($\Psi(T)$ and $\Phi(X)$) and parameters used in model training (losses, gradients, and weights for three tasks). There are tow modules, i.e. Pareto-Optimal Estimation and Pareto-Optimal Policy Learning. The data flow within a single module is represented by black thin black arrows while the data exchange between two modules is depicted by thick green arrows.}
\label{fig:framework}
\end{figure*}

Elaborating further, Pareto-Optimal Estimation operates on the principle of optimality in a multi-objective context. By situating the problem within the framework of Pareto optimality, we aim to address the inherent complexity of multi-dimensional decision-making. This involves creating a balanced and efficient policy that can navigate the delicate interplay between immediate and delayed benefits of treatments. The Pareto frontier, in this scenario, acts as a guide, delineating the set of all possible optimal solutions where each point represents a unique trade-off between short-term and long-term outcomes. Our methodology, therefore, does not just seek to identify a singular optimal treatment but aims to provide a spectrum of viable options, each calibrated to different preferences and priorities regarding short-term gains and long-term benefits.

\section{Methodology}

Guided by the above preliminary, we propose a united Pareto-Efficient Architecture (Fig.~\ref{fig:framework}), combining two main sub-module tasks: (1) Pareto-optimal Estimator for counterfactual prediction of short-term and long-term outcomes; (2) Pareto-optimal Policy Learning for trade-off between the two potential outcomes.

Specifically, in Pareto-Optimal Estimation, we first explore mutual information to learn balanced representations in contexts involving continuous treatments. Subsequently, we will investigate how the information derived from both short-term and long-term outcomes can further refine and enhance balancing representations. To avoid the issues posed by multi-task conflicts, including balancing representations and regression of short- and long-term outcomes, we introduce a novel Causal Pareto Estimator to learn the optimal networks for estimating potential outcomes, effectively navigating the complexities of these multi-dimensional tasks. Furthermore, in Pareto-Optimal Policy Learning, we reformulate the identification of the optimal treatment, specifically in terms of balancing short-term and long-term outcomes, as a Pareto optimal problem, necessitating the learned policy to find an optimal solution on the Pareto frontier while maximizing the reward.

\subsection{Pareto-Optimal Estimation}

As shown in the up-panel of Fig.~\ref{fig:framework}, given the observational data $\mathbb{D}=\{X_i, T_i, S_i, Y_i\}_{i=1}^n$, we would use representation networks  to learn $\Psi(T)$ and $\Phi(X)$, and then enforce the representation $\Phi(X)$ to capture the information of $X$ that independent with $T$, and then use the continuous Pareto technique to regress the short- and long-term outcomes with $\hat{S}=h_s(\Psi(T) \oplus \Phi(X))$ and $\hat{Y}=h_y(\Psi(T) \oplus \Phi(X), \hat{S})$. Next, we would introduce each single module in the Pareto-Optimal Estimation.

\subsubsection{Confounder Balancing for Continuous Treatment} 
\label{sec:confounderBal}

Based on the representation learning, we use the representation networks to learn $\Psi(T)$ and $\Phi(X)$, and then concatenate them into a vector $(\Psi(T) \oplus \Phi(X))$ to regress the counterfactual outcomes of short- and long-term causal effects. However, the direct information $\Phi(X)$ of confounders $X$ would confound the causal relationship between treatments and outcomes, i.e. $\Phi(X) \perp T$. 
To mitigate confounding bias caused by imbalanced covariates, the representation work \cite{CFR} suggests learning a treatment-independent balanced representation. This involves minimizing the Integral Probability Metric (IPM) distance between the treated and control groups, a technique specifically designed for binary treatment scenarios.

To address the confounding bias in continuous treatment cases, inspired by AutoIV~\cite{autoIV} and DeR-CFR~\cite{DeR-CFR}, we propose to use mutual information to measure the relevance between learned covariates representation and treatments~\cite{cheng2020club}. In mutual information estimation, we would adopt a two-phase alternative training strategy to learn the variational distribution and then estimate the mutual information.

Firstly, we would fix the parameters of representation $\Phi(X)$, and use it to approach the mean ($\mu = \mu_{\theta}(\Phi(X))$) and variance (\text{Var} = $\text{Var}_{\theta}(\Phi(X))$) of the variational distribution $q_{\theta}(\Phi(X))$, where $\mu_{\theta}(\cdot)$ and $\text{Var}_{\theta}(\cdot)$ are two learnable networks.
Then, with the representation $\Phi(X)$ fixed, we minimize the log-likelihood to optimize networks $\mu_{\theta}(\cdot)$ and $\text{Var}_{\theta}(\cdot)$ to learn the variational approximation $q_{\theta}(T|\Phi(X))$:
\begin{align}\label{equ-zx-lld}
    \mathcal{L}_{LLD}=-\frac{1}{n}\sum_{i=1}^n{\log{q_{\theta}({t}_i|\Phi({x}_i))}}, 
\end{align}
\begin{align}
    \log{q_{\theta}({t}_i|\Phi({x}_i))}=
\frac{\mu_{\theta}(\Phi({x}_i)) -t_i}{\exp (\log \text{Var}_{\theta}(\Phi({x}_i)))}  - \log \text{Var}_{\theta}(\Phi({x}_i)),
\end{align}
where $\mathcal{L}_{LLD}$ means the log likelihood function as an approximation function for probability distributions.
To reduce the relevance between the representations and the treatment, we minimize the mutual information between them. In this phase, we fix the parameters of networks $\mu_{\theta}(\cdot)$ and $\text{Var}_{\theta}(\cdot)$, and then update the representation network $\Phi(X)$ to minimize the mutual information:
\begin{align}\label{eq:mi}
    \mathcal{L}_{MI} & = \frac{1}{n^2}\sum_{i=1}^n\sum_{j=1}^n(\log{q_{\theta}(t_{i}|\Phi(x_i))}-
    \log{q_{\theta}(t_{j}|\Phi(x_i))}),
\end{align}
where $\log{q_{\theta}(t_{i}|\Phi(x_i)})$ represents the conditional log-likelihood of the positive sample pair $(\Phi(x_{i}),t_{i})$ and $q_{\theta}(t_{j}|\Phi(x_i)_{i\neq j}$ represents the negative sample pair $(\Phi(x_{i}),t_{j})_{i\neq j}$. We minimize Eq. (\ref{eq:mi}) to optimize the independent balanced representations $\Phi(X)$ via minimizing differences between the positive and negative sample pairs. Then we use them to achieve unbiased treatment effect estimation, however, this comes with the trade-off of significant information loss or increased costs due to the stringent constraints involved.
Thus, we expect to combine information from both short-term and long-term outcomes to supplement and enrich the information necessary for enhancing the shared representation framework.

\subsubsection{Shared representation for predicting both short-term and long-term outcomes}
As shown in the up-panel of Fig.~\ref{fig:framework}, once we have obtained the concatenation representation $\Psi(T) \oplus \Phi(X)$ with constraints $\Phi(X) \perp T$, we would use it as a shared representation to combine the additional information from short-term and long-term outcomes. By the way, the treatment representation function, denoted as $\Psi(T)$ and embedded in the concatenation $\Psi(T) \oplus \Phi(X)$, should be an invertible function, i.e. $\mathbb{P}(S,Y \mid \Phi(X), T) = \mathbb{P}(S,Y \mid \Psi(T) \oplus \Phi(X))$. We adopt this approach because the information from a single-dimensional treatment $T$ might get overshadowed or lost within the higher-dimensional space of $X$. To prevent this, we can use methods like repeating the treatment in the vector or applying some invertible nonlinear transformations, ensuring that the information from $T$ is preserved and not lost during the regression process. 

Then, we use a hypothesis network $h_s:\boldsymbol{\phi}\times\boldsymbol{\psi}\rightarrow s\subset\mathbb{R}$ to predict the short-term causal effect, followed by another hypothesis network $h_y:\boldsymbol{\phi}\times\boldsymbol{\psi}\times s\rightarrow y\subset\mathbb{R}$ to estimate the long-term potential outcome using this shared concatenation representation $\Psi(T) \oplus \Phi(X)$. The formal definitions of these estimands are respectively given below:
\begin{equation}\label{eq:estimation}
    \hat{s}_i=h_s(\Phi(x_i),\Psi(t_i)),\ \hat{y}_i=h_y(\Phi(x_i),\Psi(t_i),\hat{s}_i), 
\end{equation}
We aim to minimize the mean square error (MSE):
\begin{equation}\label{eq:mse-s}
    \mathcal{L}_s=\frac{1}{n} \sum_{i=1}^{n} (s_i - \hat{s}_i)^2, \quad
    \mathcal{L}_y=\frac{1}{n} \sum_{i=1}^{n} (y_i - \hat{y}_i)^2 ,
\end{equation}
where $s_i$ and $y_i$ are the true outcomes in the observational data. Then, we can integrate these loss functions into the total objective function with three hyper-parameters $\boldsymbol{w} = \{\alpha, \beta, \gamma\}$:
\begin{equation}
\label{eq:objective}
\mathcal{L} = \alpha * \mathcal{L}_{MI} + \beta * \mathcal{L}_{s} + \gamma * \mathcal{L}_{y}, 
\end{equation}
Similar to the optimization used in previous confounder balancing methods, we employ hyper-parameters to manage the trade-off among multiple tasks. However, these tasks may conflict with each other due to conflicting optimization directions. Consequently, we transform this challenge into a Pareto estimation problem to learn the optimal representation with optimal hyper-parameters \cite{zitzler1999multi}, \cite{lv2021paretoDA}, and \cite{momma2022multi}.

\begin{algorithm}[t]
      \caption{Pareto-Optimal Estimation}\label{alg:poe}
      \begin{algorithmic}[1]
        \Input Initial parameters $\xi$, step size $\eta$, \\ maximum iteration number $K$
        \FOR{$i\gets1$ \textbf{to} $K$}
        \STATE Calculate the losses $\mathcal{L}_{MI}$, $\mathcal{L}_s$, and $\mathcal{L}_y$ \\ by Eq. (\ref{eq:mi}) and Eq. (\ref{eq:mse-s})
            \STATE Calculate $[g_1,g_2,g_y]$ $\gets[\nabla_\xi\mathcal{L}_{MI},\nabla_\xi\mathcal{L}_s,\nabla_\xi\mathcal{L}_y]$
            \STATE Obtain $\boldsymbol{w}$ by Eq. (\ref{eq:optimal})
            \STATE Calculate optimization direction  by $\boldsymbol{d}\gets\boldsymbol{G}\boldsymbol{w}$
            \STATE Update parameters by $\xi^{i+1}\gets\xi^i-\eta\boldsymbol{d}$
        \ENDFOR
        \Output Updated parameters $\xi^{K}$
      \end{algorithmic}
    \end{algorithm}

\subsubsection{Pareto Optimization}
Inspired by previous Pareto optimization works~\cite{lv2021paretoDA,ma2020efficient,lin2019pareto}, we would like to transfer the trade-off among multiple tasks into a continuous Pareto Optimization Problem. Firstly, we would like to figure out the gradients of three components in the objective function (Eq. (\ref{eq:objective})) as $\boldsymbol{G}= (g_1,g_2,g_3)=(\nabla\mathcal{L}_{MI},\nabla\mathcal{L}_s,\nabla\mathcal{L}_y)$ and their corresponding trade-off hyper-parameters as $\boldsymbol{w} = \{w_1,w_2,w_3\}=\{\alpha, \beta, \gamma\} \in \mathbb{R}_{+}^3$. Considering that the scopes of each loss are various, we employ padding operator to make them consistent with each other.

Inspired by the objective function that is suggested in MGDA~\cite{sener2018multi}, we consider
\begin{equation}
    \begin{split}
    &\min\limits_{\boldsymbol{w}}\ J=\frac{1}{2}\left\|\sum_{i=1}^mw_i\nabla g_i(\xi)\right\|_2^2 = \frac{1}{2}\boldsymbol{w}^T\nabla g(\xi)(\nabla g(\xi))^T\boldsymbol{w} \\
    &s.t.\ \boldsymbol{w}^T\boldsymbol{1}_m=1,\ \boldsymbol{w}\geq 0,
    \end{split}
\end{equation}
where $m$ is the number of multiple tasks ($m=3$ in POE), and $\boldsymbol{w}$ represents the weights to adaptively trade-off among them. Moreover, each $g_i(\cdot)$ refers to one learning objective (i.e. gradient function for each loss). Note that $\xi$ is the parameters of the entire module, including $\Psi(\cdot)$, $\Phi(\cdot)$, $h_s(\cdot)$, and $h_y(\cdot)$ in Fig.~\ref{fig:framework}.
We can rewrite it as an augmented Lagrangian form:
\begin{equation}\small\label{eq:pareto}
    J=\frac{1}{2}\boldsymbol{w}^T\nabla g(\xi)(\nabla g(\xi))^T\boldsymbol{w}+\mu(\boldsymbol{w}^T\boldsymbol{1}_m-1)+\frac{\rho}{2}\|\boldsymbol{w}^T\boldsymbol{1}_m-1\|_2^2,
\end{equation}
where $\mu$ and $\rho$ are the Lagrangian coefficient and augmented Lagrangian coefficient, respectively. Therefore, the optimization process can be expressed as
\begin{equation}\label{eq:optimal}
    \left\{
\begin{aligned}
  &\ \boldsymbol{w}=\max\left(0,\left(\nabla g(\xi)(\nabla g(\xi))^T+\rho\boldsymbol{I}\right)^{-1}(\rho\boldsymbol{I}-\mu\boldsymbol{I})\right), \\
  &\ \mu\gets\mu+\rho(\boldsymbol{w}^T\boldsymbol{1}_m-1).
\end{aligned}
\right.
\end{equation}

The update direction in each optimization step can be regarded as $\boldsymbol{d}=\boldsymbol{G}\boldsymbol{w}$. Shared parameters $\xi^i$ in the $i$-th step can be updated by $\xi^{i+1}=\xi^i-\eta\boldsymbol{d}^*$, where $\eta$ is the step size. We summarize this process in Algorithm~\ref{alg:poe}. Using the learned optimal $\boldsymbol{w}^*$, we can achieve optimal balanced representation for short-term and long-term outcomes regression.

\begin{algorithm}[t]
      \caption{Pareto-Optimal Policy Learning}\label{alg:popl}
      \begin{algorithmic}[1]
        \Input Initial parameters $\zeta$, step size $\lambda$, \\ maximum iteration number $N$
        \FOR{$i\gets1$ \textbf{to} $N$}
            \STATE Calculate the losses $\mathcal{R}_{s}$ and $\mathcal{R}_y$ \\ by Eq. (\ref{eq:regret_s}) and Eq. (\ref{eq:regret_y})
            \STATE Calculate $[g_4,g_5]$ $\gets[\nabla_\zeta\mathcal{R}_{s},\nabla_\zeta\mathcal{R}_y]$
            \STATE Obtain $\boldsymbol{w}$ by Eq. (\ref{eq:optimal})
            \STATE Calculate optimization direction  by $\boldsymbol{d}\gets\boldsymbol{G}\boldsymbol{w}$
            \STATE Update parameters by $\zeta^{i+1}\gets\zeta^i-\lambda\boldsymbol{d}$
        \ENDFOR
        \Output Updated parameters $\zeta^{N}$
      \end{algorithmic}
    \end{algorithm}

\subsection{Pareto-Optimal Policy Learning}~\label{sec:policy}
In previous section, we introduced a novel Pareto-Optimal Estimation module, crafted to accurately estimate potential short-term and long-term outcomes for any specific manipulated intervention. Consequently, a direct motivation emerges: to discover an optimal policy that assists practitioners in identifying the most effective treatment for maximizing the reward. 

In the real-world application, the observations typically are only the pre-treatment variable covariates $X$, thus, the objective of the optimal policy is to explore which values of $T$ would lead to Pareto optimal solutions of the short-term and long-term outcomes. Similar to the Pareto estimation module described above, we train a deterministic policy backbone $\Pi(X):\mathcal{X}\rightarrow\mathcal{T}\subset\mathbb{R}$. With the estimator backbone, potential outcomes of the decided policy $\Pi(X)$ can be calculated. In order to train the policy backbone, its objective is to minimize the regret loss of each potential outcome, which is defined as the difference between the maximum outcome minus the expected outcome of $\Pi(X)$. As for the short-term outcome, the regret loss can be expressed as:
\begin{equation}\label{eq:regret_s}
    \mathcal{R}_s=\frac{1}{N}\sum_{i=1}^N\max_{t_i} \{\hat{s}_i - h_s(\Phi(x_i),\Psi(\Pi(x_i)))\}.
\end{equation}
Similarly, the objective with respect to the long-term outcome is given as follows.
\begin{equation}\label{eq:regret_y}
    \begin{split}
        \mathcal{R}_y=\frac{1}{N}\sum_{i=1}^N\max_{t_i} \{\hat{y}_i - h_y(\Phi(x_i),\Psi(\Pi(x_i)),\hat{h}_s)\}.
    \end{split}
\end{equation}
We also utilize the continuous Pareto optimization algorithm to trade-off between the two regrets and further update this backbone. Details are described in Algorithm~\ref{alg:popl}, whose aim is to update the Pareto set in each step.

\subsection{Overall}
We conclude the overall workflow in Algorithm~\ref{alg:workflow}. In Pareto-Optimal Estimation, networks $\Psi(\cdot)$ and $\Phi(\cdot)$ are trained to learn the representation of $T$ and $X$, respectively. Afterwards, they are inputted into hypothesis network $h_s(\cdot)$ to predict the short-term potential outcome, and we denote the predicted results as $\hat{S}$. In a similar way, $\hat{Y}$ is output by another network $h_y(\cdot)$ while $\hat{S}$ serves as an additional input for it. Regression loss $\mathcal{L}_s$ between $\hat{S}$ and $S$ is calculated, together with $\mathcal{L}_y$ between $\hat{Y}$ and $Y$. The loss $\mathcal{L}_{MI}$ uses $\Phi(X)$ and $T$ to measure the effectiveness of representation learning. After calculating the gradients ($[\boldsymbol{g}_1,\boldsymbol{g}_2,\boldsymbol{g}_3]$) of all losses, a Pareto optimization algorithm is applied to trade-off among these conflicting objectives, adaptively adjusting their weights ($[\boldsymbol{w}_1,\boldsymbol{w}_2,\boldsymbol{w}_3]$) to determine the optimization direction for model updates.

As for Pareto-Optimal Policy Learning, given $X$, network $\Pi(\cdot)$ is trained to predict the optimal treatment value $\hat{T}$ of which the short- and long-term outcomes are Pareto optimal. Afterwards, $X$ and $\hat{T}$ are input into the estimator (depicted as the green arrow on the left) to output the corresponding $\hat{S}$ and $\hat{Y}$ (demonstrated by the green arrow on the right). Two regret losses ($\mathcal{R}_s$ and $\mathcal{R}_y$) are then obtained. Similar Pareto optimization strategy is applied here to update $\Pi(\cdot)$.

\begin{algorithm}[t]
      \caption{Overall Workflow}\label{alg:workflow}
      \begin{algorithmic}[1]
        \Input Dataset $\mathbb{D}$, initial parameters of POE $\xi$, initial parameters of POPL $\zeta$, step size $\eta$, maximum iteration $K$
        \FOR{$i\gets1$ \textbf{to} $K$}
            \STATE inputs $\gets[\Phi(\mathbb{D}.X),\Psi(\mathbb{D}.T)]$
            \STATE $\hat{S}\gets h_s(inputs)$\ \ \ \ \ \ \ \ \ \ \ \ \ \ \ \ \ \ \ \ // Eq. (\ref{eq:estimation})
            \STATE $\hat{Y}\gets h_s(inputs, \hat{S})$\ \ \ \ \ \ \ \ \ \ \ \ \ \ \ \ // Eq. (\ref{eq:estimation})
            \STATE $\mathcal{L}_{MI}\gets MI(\Phi(X),T)$\ \ \ \ \ \ \ \ \ \ \ \ // Eq. (\ref{eq:mi})
            \STATE $\mathcal{L}_s\gets MSE(\hat{S},\mathbb{D}.S)$\ \ \ \ \ \ \ \ \ \ \ \ \ \ // Eq. (\ref{eq:mse-s})
            \STATE $\mathcal{L}_y\gets MSE(\hat{Y},\mathbb{D}.Y)$\ \ \ \ \ \ \ \ \ \ \ \ \ // Eq. (\ref{eq:mse-s})
            \STATE $\xi\gets\xi'$ after $\mathcal{L}$.backward()
        \ENDFOR
        \STATE $\xi\gets$ POE.train($\xi$, $\eta$, $K$)\ \ \ \ \ \ \ \ \ \ \ \ \ \ // Algorithm~\ref{alg:poe}
        \STATE $[s^*,y^*]\gets[\max(\mathbb{D}.S),\max(\mathbb{D}.Y)]$ 
        \FOR{$i\gets1$ \textbf{to} $K$}
            \STATE $\hat{T}\gets\Pi(\mathbb{D}.X)$
            \STATE $\hat{S}\gets h_s(\Phi(\mathbb{D}.X),\hat{T})$\ \ \ \ \ \ \ \ \ \ \ \ \ \ \ // Eq. (\ref{eq:estimation})
            \STATE $\hat{Y}\gets h_y(\Phi(\mathbb{D}.X),\hat{T},\hat{S})$\ \ \ \ \ \ \ \ \ \ \ // Eq. (\ref{eq:estimation})
            \STATE $\mathcal{R}_s\gets Regret(\hat{S},s^*)$\ \ \ \ \ \ \ \ \ \ \ \ \ \ // Eq. (\ref{eq:regret_s})
            \STATE $\mathcal{R}_y\gets Regret(\hat{Y},y^*)$\ \ \ \ \ \ \ \ \ \ \ \ \ \ // Eq. (\ref{eq:regret_y})
            \STATE $\zeta\gets\eta'$ after $\mathcal{R}$.backward()
        \ENDFOR
        \STATE $\zeta\gets$ POPL.train($\zeta$, $\eta$, $K$)\ \ \ \ \ \ \ \ \ \ \ \ // Algorithm~\ref{alg:popl}
        \Output Updated parameters $\xi,\zeta$
      \end{algorithmic}
    \end{algorithm}

\section{Experiments}
In this section, we first introduce the five datasets and some detailed constructions of the synthetic ones in them. Afterwards, we conduct extensive experiments to evaluate the performance of our proposed model.

\subsection{Datasets}~\label{sec:dataset}
There are totally five datasets we use in the evaluation, including one real-life dataset (Crime), three semi-synthetic datasets (IHDP, Jobs, and Twins), and one simulation dataset (Simulation). We summarize the statistics of the five datasets in Table~\ref{tab:dataset}. Generally speaking, it is more challenging to make counterfactual predictions if there is a smaller number of samples in the training set.

\textbf{Crime}\footnote{\url{https://tandf.figshare.com/ndownloader/articles/24262352/versions/1}}. Some researchers have studied the impact of New York’s bail reform~\cite{data-crime}, which is implemented on January $1$, $2020$. The raw data records the aggregate level of a specific crime in $27$ cities for each day from January $1$, $2018$ to March $15$, $2020$. We convert the crime data for $805$ days into monthly average crime rates (normalized after dividing by the population). We use the data of all the months before January $1$, $2020$ as the corivates. Including the city names and crime categories (two-level classification), we obtain a total of $27$ covariates. The average crime rates of January and March in $2020$ are treated as the short-term and long-term outcomes, respectively. Note that the treatment in this dataset is binary, i.e only the crimes in New York belong to the treated group and the rest consist of the control group.

\begin{table}[t]
\centering
\caption{Statistics of datasets.}
\label{tab:dataset}
\begin{tabular}{>{\centering\arraybackslash}m{1.5cm}*{2}{>{\centering\arraybackslash}m{1cm}}{>{\centering\arraybackslash}m{2cm}}}
\toprule
dataset & \# train & \# test & dimension of $X$ \\
\midrule
Crime & $171$ & $43$ & $27$ \\
IHDP & $537$ & $135$ & $25$ \\
Jobs & $2,056$ & $514$ & $17$ \\
Twins & $3,795$ & $949$ & $38$ \\
Simulation & $16,000$ & $4,000$ & $2$ \\
\bottomrule
\end{tabular}
\end{table}

\textbf{IHDP}\footnote{\url{https://www.fredjo.com/}}. It is a commonly adopted benchmark dataset that is collected from Infant Health and Development Program~\cite{IHDP}. It is a longitudinal research conducted in the United States from $1985$ to $1993$, with the purpose to study the effect of low-birthweight and premature brith on the infants’ future development (measured by cognitive test score). There are $25$ covariates covering various aspects of the infants together with their mothers, such as neonatal health index, prenatal care, mother’s age, education, etc. We utilize these covariates, denoted as $X$, to generate the continuous treatments by 
\begin{equation}
    T = \sum_{i=1}^{25} \cos(1+X_i^2).
\end{equation}
The short-term and long-term outcomes are designed as
\begin{equation}
    \left\{
\begin{aligned}
  &\ S=2.5\sin(2+T) + 0.25 \sum_{i=1}^{25} e^{-X_i^2} + 1.25, \\
  &\ Y=0.1T^2-\log(S)+2\sum_{i=1}^{25}X_i+5.
\end{aligned}
\right.
\end{equation}

\textbf{Jobs}. This dataset~\cite{jobs} comprises of the data from two sources, i.e. Lalonde experiment and the Panel Study of Income Dynamics (PSID). The download link is consistent with the IHDP dataset, which is available in the footnote. This study is focused on how the job training could affect an individual's employment status. Information such as age, education, ethnicity, as well as previous earnings are included in $X$. We use the $17$ covariates to generate the treatment variables by the following equation:
\begin{equation}
    T=0.2\sum_{i=1}^{17}\left(\sin(X_i)+e^{-X_i^2}\right).
\end{equation}
Similar to the generation formulations utilized in IHDP, the two outcome variables in Jobs dataset are defined as 
\begin{equation}
    \left\{
\begin{aligned}
  &\ S=1.7\sin(2T)+0.05\sum_{i=1}^{17}X_i+3.4, \\
  &\ Y=0.7T-S+0.02\log(1+X_i^2)+5.
\end{aligned}
\right.
\end{equation}

\textbf{Twins}\footnote{\url{https://www.nber.org/research/data/linked-birthinfant-death-cohort-data}}. It is collected from all births in the USA between $1989$-$1991$, and only the twins weighing less than $2$kg are recorded without missing features~\cite{data-twins}. Covariates measure information in $38$ dimensions, including pregnancy, the quality of care, pregnancy risk factors, residence, etc. Birth weight is regarded as the treatment ($T=1$ for the heavier one in the twin and $T=0$ for the other). The outcome of interest is $1$-year mortality. In our experiment, we generate the treatment with the help of $38$ covariates through 
\begin{equation}
    T=0.5\sum_{i=1}^{38}\log\left(1+e^{X_i}\right)-15.
\end{equation}
The expressions of short-term and long-term outcomes are given as follows.
\begin{equation}
    \left\{
\begin{aligned}
  &\ S=0.75\sin T+0.02\sum_{i=1}^{38}e^{-X_i^2}+2, \\
  &\ Y=0.2e^{\sqrt{T}}-0.2\cos S+0.001\sum_{i=1}^{38}X_i^2+2.
\end{aligned}
\right.
\end{equation}

\textbf{Simulation}. Considering the diversity in the number of covariates and samples, we also generate a simulated dataset of $20,000$ samples with $2$ covarites. Each dimension of the covarites follows a uniform distribution, i.e $x_i\sim U(0,2)$. Treatments are assigned by 
\begin{equation}
    T=\sum_{i=1}^2\log(1+e^{X_i}).
\end{equation}
The short-term and long-term outcomes are defined as 
\begin{equation}
    \left\{
\begin{aligned}
  &\ S=0.4\sin T+0.2\sum_{i=1}^2e^{-X_i^2}+1, \\
  &\ Y=0.1e^{\sqrt{T}}-0.1\cos S+0.01\sum_{i=1}^2X_i^2+1.
\end{aligned}
\right.
\end{equation}

Note that covariates are quite different among the three semi-synthetic datasets and the simulated dataset, and we apply different equations to generate treatments and outcomes so as to guarantee the conflicts between $S$ and $Y$.

\begin{table*}[t]
\centering
\caption{Estimation results of semi-synthetic datasets (IHDP, Jobs, Twins).}
\label{tab:cf-semi}
\begin{tabular}{>{\centering\arraybackslash}m{1.5cm}*{6}{>{\centering\arraybackslash}m{2.1cm}}}
\toprule
 & \multicolumn{2}{c}{\textbf{IHDP}} & \multicolumn{2}{c}{\textbf{Jobs}} & \multicolumn{2}{c}{\textbf{Twins}} \\
 & $\mathbf{MSE_s}$ & $\mathbf{MSE_y}$ & $\mathbf{MSE_s}$ & $\mathbf{MSE_y}$ & $\mathbf{MSE_s}$ & $\mathbf{MSE_y}$  \\
\midrule
TarNet & $1.434\pm 0.031$ & $0.297\pm 0.051$ & $1.308\pm 0.048$ & $8.561\pm 1.279$ & $0.583\pm 0.024$ & $0.348\pm 0.014$ \\
DRNet & $3.284\pm 0.668$ & $0.585\pm 0.182$ & $1.874\pm 0.114$ & $7.633\pm 1.249$ & $0.411\pm 0.203$ & $0.286\pm 0.137$ \\
VCNet & $1.636\pm 0.060$ & $0.397\pm 0.111$ & $1.754\pm 0.023$ & $6.881\pm 0.606$ & $0.177\pm 0.112$ & $0.155\pm 0.063$ \\
CFR & $0.461\pm 0.050$ & $0.243\pm 0.102$ & $1.001\pm 0.318$ & $1.038\pm 0.349$ & $0.071\pm 0.023$ & $0.036\pm 0.018$ \\ \hline
\textbf{Ours} & $\textbf{0.382}\boldsymbol{\pm} \textbf{0.025}$ & $\textbf{0.182}\boldsymbol{\pm} \textbf{0.024}$ & $\textbf{0.238}\boldsymbol{\pm} \textbf{0.096}$ & $\textbf{0.185}\boldsymbol{\pm} \textbf{0.063}$ & $\textbf{0.041}\boldsymbol{\pm} \textbf{0.001}$ & $\textbf{0.015}\boldsymbol{\pm} \textbf{0.001}$ \\
\bottomrule
\end{tabular}
\end{table*}

\subsection{Baselines}\label{sec:baseline}
We apply the models with similar architecture as the baselines, including Treatment-Agnostic Representation Network (TarNet)~\cite{CFR}, Counterfactual Regression (CFR)~\cite{CFR}, Dose Response Network (DRNet)~\cite{DRNet}, and Varying Coefficient Neural Network (VCNet)~\cite{VCNet}.

\textbf{TARNet}. It is a classic model applied in the binary treatment setting. There is a representation network to learn a shared embedding of $X$ across the treated group and the control group, together with two hypothesis networks to predict the outcomes of each group. This is because the dimension of $X$ is often high while that of $T$ is only $1$, thus weakening the effect of $T$ in counterfactual prediction. By separately learning for each value of $T$, its impact can be highlighted and distinguished from that of $X$.

\textbf{CFR}. There is an additional module in CFR compared to TARNet, which utilizes Integral Probability Metrics (IPM) to make sure the covariate distribution balance between different groups. This is a further effort made to eliminate the causal path from confounders to the treatment.

\textbf{DRNet}. It can be seen as an extended version of TARNet that makes attempt to estimate the effect of continuous treatment. There is also a base layer to learn the representation of $X$, and several treatment layers to learn specific representations of each kind of $T$. For each treatment layer, multiple heads are nested to divide the domain of treatment, i.e. $\mathcal{T}$, into multiple intervals and learn the corresponding dosage function, thereby obtaining the outcome. Strictly speaking, DRNet is not continuous but segmented regression. Performance of DRNet becomes poor when there are abrupt changes of estimation especially at the boundary points of dosage intervals.

\textbf{VCNet}. The representation of covariates $X$ is extracted as $Z$ through a non-linear mapping, where a propensity estimation in Dragonnet~\cite{Dragonnet} is applied to ensure the distribution balance among various treatment values. Furthermore, there is only one varying coefficient prediction head with the purpose of counterfactual prediction given $T$ and $Z$.

\begin{table}[t]
\centering
\caption{Estimation results of Simulation dataset.}
\label{tab:cf-sim}
\begin{tabular}{>{\centering\arraybackslash}m{1.5cm}*{2}{>{\centering\arraybackslash}m{2.5cm}}}
\toprule
 & $\mathbf{MSE_s}$ & $\mathbf{MSE_y}$ \\
\midrule
TarNet & $0.442\pm 0.028$ & $0.206\pm 0.019$ \\
DRNet & $0.515\pm 0.093$ & $0.216\pm 0.029$ \\
VCNet & $0.293\pm 0.017$ & $0.184\pm 0.008$ \\
CFR & $0.222\pm 0.211$ & $0.088\pm 0.134$ \\ \hline
\textbf{Ours} & $\textbf{0.012}\boldsymbol{\pm} \textbf{0.014} $ & $\textbf{0.005}\boldsymbol{\pm} \textbf{0.006}$ \\
\bottomrule
\end{tabular}
\end{table}

Specifically, we train two networks to separately predict the short-term outcome and long-term outcome. The codes we utilize for the first three methods are obtained from the open-source\footnote{\url{https://github.com/lushleaf/varying-coefficient-net-with-functional-tr}} of VCNet. We develop TarNet into CFR by adding a mutual information module between $T$ and the representation of $X$. The loss calculating such mutual information is exactly the representation loss designed in our model, whose formal definition is given in Eq. (\ref{eq:mi}).

\subsection{Results}
\subsubsection{Estimation}\label{sec:res-estimation}
We uniformly select $50$ points within the possible interval of $\mathcal{T}$ as $t^{cf}$ for counterfactual predictions, and use Mean Squared Error (MSE) to evaluate the ability of all the models that are mentioned above. After training the models with $10$ randomly picked seeds, all the results are reported in the form of $(\text{mean value}\pm\text{standard deviation})$.

Experiment results on three semi-synthetic datasets are concluded in Table~\ref{tab:cf-semi}, providing a clear comparison between our method and the four baselines in terms of estimation performance. Our method consistently outperforms the four baselines across all datasets. On the IHDP dataset, although the CFR method shows competitive performance with relatively low $\mathbf{MSE_s}$ ($0.461\pm0.050$) and $\mathbf{MSE_y}$ ($0.243\pm0.102$), our method still achieves the lowest $\mathbf{MSE_s}$ ($0.382\pm0.025$) and $\mathbf{MSE_y}$ ($0.182\pm0.024$). In the case of Jobs dataset, our method surpasses the second-best approach by a significant margin of $0.763$ in $\mathbf{MSE_s}$ and $0.853$ in $\mathbf{MSE_y}$, indicating our model's superiority in terms of accuracy and stability. The Twins dataset is less challenging due to the sufficient number of samples for training. Although all the baselines perform well on this dataset, our approach still achieves the best performance with $\mathbf{MSE_s}=0.041$ and $\mathbf{MSE_y}=0.015$.

We also conduct experiments on a simulated dataset, the results of which are demonstrated in Table~\ref{tab:cf-sim}. Due to the large size of training data, each model has been extensively trained, resulting in relatively low values of $\mathbf{MSE_s}$ and $\mathbf{MSE_y}$. According to this table, TARNet and DRNet exhibit similar performance, while VCNet achieves further improvement over them. CFR remains the closest approach to our proposed method, particularly demonstrating strong performance in terms of $\mathbf{MSE_y}$ ($0.088\pm0.134$). Even in this situation, our proposed method still achieves significant improvement (an order of magnitude) compared to CFR, again validating the superiority of our approach.

\begin{table}[t]
\centering
\caption{Estimation results of Crime dataset.}
\label{tab:cf-crime}
\begin{tabular}{>{\centering\arraybackslash}m{1.5cm}*{2}{>{\centering\arraybackslash}m{2.5cm}}}
\toprule
 & $\mathbf{MSE_s}$ & $\mathbf{MSE_y}$ \\
\midrule
TarNet & $4.322\pm 2.494$ & $5.221\pm 1.792$ \\
DRNet & $5.387\pm 1.104$ & $3.256\pm 1.459$ \\
VCNet & $1.155\pm 0.498$ & $2.639\pm 1.511$ \\
CFR & $0.911\pm 0.296$ & $1.346\pm 0.290$ \\ \hline
\textbf{Ours} & $\textbf{0.838}\boldsymbol{\pm} \textbf{0.149}$ & $\textbf{0.510}\boldsymbol{\pm} \textbf{0.078}$ \\
\bottomrule
\end{tabular}
\end{table}

\begin{figure*}[t]
\centering
\includegraphics[width=1\textwidth]{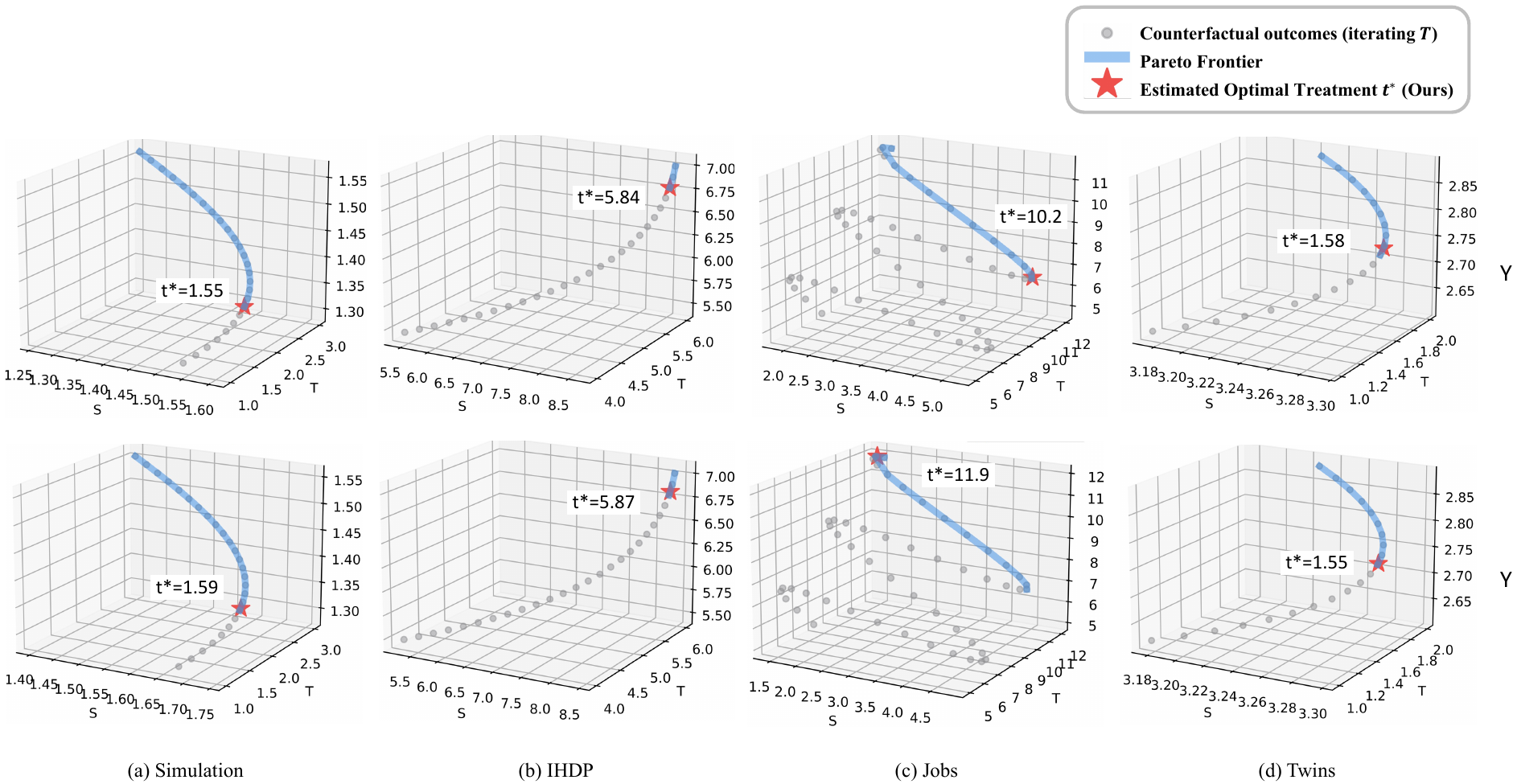}
\caption{Visualization of policy learning, where the optimal value $t^*$ estimated by our model is always located at the Pareto frontier. The x-axis represents the value of short-term outcome $S$, y-axis referring to $T$ and z-axis is long-term outcome $Y$. This figure depicts all potential outcomes, including $S$ and $Y$, for $t\in[1,3]$, $[4,6]$, $[5,12]$, and $[1,2]$ on Simulation, IHDP, Jobs, and Twins dataset, respectively. Crime dataset is not used here due to the binary treatment setting and lack of groundtruth. We choose to demonstrate the experimental results using three-dimensional graphics to provide a clearer illustration of the conflicts between $S$ and $Y$ as $T$ varies.}
\label{fig:vis}
\end{figure*}

The evaluation on the real-life dataset can offer a comprehensive insight into the performance of the methods, and we report the counterfactual prediction results on the Crime dataset in Table~\ref{tab:cf-crime}. Note that the treatment is binary here, and only the crimes in New York City are regarded as the treated group (implementing bail reform). For each sample in the treated group, we search for the three most similar samples with the highest Pearson correlation coefficients from the control group. The formal definition of correlation coefficient between unit $i$ and $j$ is
\begin{equation}
    \rho(X_i,X_j)=\frac{\text{cov}(X_i,X_j)}{\sigma (X_i)\sigma (X_j)},
\end{equation}
where $\rho(X_i,X_j)$ means their covariance, and $\sigma (X_i)$ and $\sigma (X_j)$ are the standard deviations. Afterward, we calculate a weighted average of the selected samples' outcomes as the groundtruth for counterfactual prediction. Based on the reported results, we can see that VCNet exhibits improved performance compared to TarNet and DRNet, and CFR achieves the lowest MSE values of all the baselines. In comparison with CFR, although our method only achieves a decrease of $0.073$ in $\mathbf{MSE_s}$, it still stands out with a remarkable improvement of $0.836$ in $\mathbf{MSE_y}$.

In summary, our model outperforms all the baselines in terms of predicting both the short-term and long-term outcomes whichever dataset is applied. The superiority of our model in $\mathbf{MSE_y}$ is easy to explain. Different from the baselines that train two separate networks to predict $S$ and $Y$ independently, we leverage the predicted value $\hat{S}$, which is derived from the first prediction head, to inform the prediction of $Y$ in the second head. By incorporating this additional information, our model achieves more accurate estimations of the causal effect on $Y$. Furthermore, our model also demonstrates superior performance in predicting short-term outcomes. This can be attributed to the fact that the long-term outcome $Y$ contains information from both the covariates $X$ and the short-term outcome $S$. Therefore, by optimizing the prediction of $Y$, our model effectively enhances the representation learning of $X$ and improves the prediction accuracy of $S$ in turn.

\begin{table*}[t]
\centering
\caption{Results of ablation study on IHDP, Jobs, Twins and Simulation datasets.}
\label{tab:ablation}
\begin{tabular}{>{\centering\arraybackslash}m{1.55cm}*{8}{>{\centering\arraybackslash}m{1.6cm}}}
\toprule
& \multicolumn{2}{c}{\textbf{IHDP}} & \multicolumn{2}{c}{\textbf{Jobs}} & \multicolumn{2}{c}{\textbf{Twins}} & \multicolumn{2}{c}{\textbf{Simulation}} \\
 & $\mathbf{MSE_s}$ & $\mathbf{MSE_y}$ & $\mathbf{MSE_s}$ & $\mathbf{MSE_y}$ & $\mathbf{MSE_s}$ & $\mathbf{MSE_y}$ & $\mathbf{MSE_s}$ & $\mathbf{MSE_y}$ \\
\midrule
CFR-$S$  & $0.461\pm 0.050$ & $-$ & $1.001\pm 0.318$ & $-$ & $0.071\pm 0.023$ & $-$ & $0.222\pm 0.211$ & $-$ \\
CFR-$Y$ & $-$ & $0.243\pm 0.102$ & $-$ & $1.038\pm 0.349$ & $-$ & $0.036\pm 0.018$ & $-$ & $0.088\pm 0.134$ \\
Joint CFR & $0.425\pm 0.020$ & $0.220\pm 0.136$ & $0.519\pm 0.192$ & $0.506\pm 0.192$ & $0.057\pm 0.025$ & $0.024\pm 0.019$ & $0.052\pm 0.053$ & $0.023\pm 0.021$ \\
+$\hat{S}$ & $0.414\pm 0.034$ & $0.195\pm 0.016$ & $0.437\pm 0.155$ & $0.417\pm 0.186$ & $0.048\pm 0.003$ & $0.018\pm 0.002$ & $0.023\pm 0.023$ & $0.022\pm 0.028$ \\ 
\textbf{+$\boldsymbol{\hat{S}}$+Pareto} &  $\textbf{0.382}\boldsymbol{\pm} \textbf{0.025}$ & $\textbf{0.182}\boldsymbol{\pm} \textbf{0.024}$ & $\textbf{0.238}\boldsymbol{\pm} \textbf{0.096}$ & $\textbf{0.185}\boldsymbol{\pm} \textbf{0.063}$ & $\textbf{0.041}\boldsymbol{\pm} \textbf{0.001}$ & $\textbf{0.015}\boldsymbol{\pm} \textbf{0.001}$ & $\textbf{0.012}\boldsymbol{\pm} \textbf{0.014}$ & $\textbf{0.005}\boldsymbol{\pm} \textbf{0.006}$ \\
\bottomrule
\end{tabular}
\end{table*}

\begin{table*}[t]
\centering
\caption{Results of hyper-parameter study on IHDP, Jobs, Twins and Simulation datasets.}
\label{tab:hyper}
\begin{tabular}{>{\centering\arraybackslash}m{1cm}*{8}{>{\centering\arraybackslash}m{1.6cm}}}
\toprule
& \multicolumn{2}{c}{\textbf{IHDP}} & \multicolumn{2}{c}{\textbf{Jobs}} & \multicolumn{2}{c}{\textbf{Twins}} & \multicolumn{2}{c}{\textbf{Simulation}} \\
 $\alpha (w_1)$ & $\mathbf{MSE_s}$ & $\mathbf{MSE_y}$ & $\mathbf{MSE_s}$ & $\mathbf{MSE_y}$ & $\mathbf{MSE_s}$ & $\mathbf{MSE_y}$ & $\mathbf{MSE_s}$ & $\mathbf{MSE_y}$ \\
\midrule
$0.1$ & $1.222\pm 0.182$ & $0.085\pm 0.015$ & $0.291\pm 0.165$ & $0.231\pm 0.127$ & $0.043\pm 0.002$ & $0.015\pm 0.001$ & $0.035\pm 0.035$ & $0.009\pm 0.005$\\
$0.3$ & $0.661\pm 0.059$ & $0.136\pm 0.034$ & $0.284\pm 0.100$ & $0.240\pm 0.063$ & $0.042\pm 0.001$ & $0.015\pm 0.001$ & $0.045\pm 0.049$ & $0.021\pm 0.036$ \\ 
$0.5$ & $0.441\pm 0.043$ & $0.188\pm 0.028$ & $0.393\pm 0.161$ & $0.341\pm 0.143$ & $0.044\pm 0.002$ & $0.016\pm 0.001$ & $0.050\pm 0.046$ & $0.021\pm 0.046$ \\
$0.7$ & $0.402\pm 0.031$ & $0.186\pm 0.032$ & $0.485\pm 0.145$ & $0.423\pm 0.152$ & $0.042\pm 0.002$ & $0.017\pm 0.001$ & $0.042\pm 0.040$ & $0.022\pm 0.033$ \\ 
$0.9$ & $0.386\pm 0.032$ & $0.196\pm 0.028$ & $0.520\pm 0.161$ & $0.485\pm 0.181$ & $0.043\pm 0.002$ & $0.018\pm 0.002$ & $0.049\pm 0.030$ & $0.018\pm 0.016$ \\ \hline
\textbf{Ours} &  $\textbf{0.382}\boldsymbol{\pm} \textbf{0.025}$ & $\textbf{0.182}\boldsymbol{\pm} \textbf{0.024}$ & $\textbf{0.238}\boldsymbol{\pm} \textbf{0.096}$ & $\textbf{0.185}\boldsymbol{\pm} \textbf{0.063}$ & $\textbf{0.041}\boldsymbol{\pm} \textbf{0.001}$ & $\textbf{0.015}\boldsymbol{\pm} \textbf{0.001}$ & $\textbf{0.012}\boldsymbol{\pm} \textbf{0.014}$ & $\textbf{0.005}\boldsymbol{\pm} \textbf{0.006}$ \\
\bottomrule
\end{tabular}
\end{table*}

\subsubsection{Policy Learning}
Visualization of the policy learning on different datasets is demonstrated in Fig.~\ref{fig:vis}, where the x-axis corresponds to short-term outcomes $S$, the y-axis represents the values of assigned treatment $T$, and the z-axis refers to long-term outcomes $Y$. Setting the coordinate axes in this way gives us a more clear insight of the conflicting relationship between $S$ (x-axis) and $Y$ (z-axis). Note that the real-life dataset is not applied because of the binary treatment setting and the lack of ground truth for evaluation.

For each dataset, we randomly select two samples as representatives and plot the three-dimensional visualization of the decision-making. Specifically, we discretize the interval $\mathcal{T}$ of treatment by uniform sampling, and regard the sampled points as possible values $t^{cf}$ for counterfactual prediction. Afterwards, we iterate over all these $t^{cf}$, and use the generation formulas (mentioned in Section~\ref{sec:dataset}) to yield the corresponding short-term outcome $s^{cf}$ and long-term outcome $y^{cf}$. These counterfactual outcomes are depicted as grey points in Fig.~\ref{fig:vis}. The main purpose is to determine the Pareto frontier between short-term and long-term outcomes for these points, and the frontier is marked in blue.
Note that larger values for both $S$ and $Y$ are preferred in our setting, meaning that the Pareto frontier is the upper-right boundary of these points.

The optimal treatment $t^*$ for a sample is predicted by the policy-learning module $\Pi(x_i)$ in Section~\ref{sec:policy}. We mark the data point $(s^*,t^*,y^*)$ as a red star in Fig.~\ref{fig:vis}. It can be seen that the red star is located on the Pareto frontier, meaning that the outcomes of $t^*$ determined by our model will not be dominated by the outcomes of any other values for $T$. The effectiveness of policy learning is validated in each dataset.

\subsubsection{Ablation Study}
We conduct the ablation study in both the simulated and semi-synthetic datasets, whose results are demonstrated in Table~\ref{tab:ablation}. The model called CFR-$S$ and CFR-$Y$ here correspond to the baseline CFR mentioned in the treatment effect estimation experiment. Data reported in these two rows are the same with those in Table~\ref{tab:cf-semi} and Table~\ref{tab:cf-sim}. We decompose it into CFR-$S$ and CFR-$Y$ primarily to facilitate a more intuitive comparison with Joint CFR. In our statement in Section~\ref{sec:res-estimation}, outcome $Y$ contains the information about $S$ and $X$. Therefore, it can facilitate mutual enhancement to some extent if multiple tasks are learned jointly. The improvements in both the $\mathbf{MSE_s}$ and $\mathbf{MSE_y}$ validate our viewpoint. Due to the sparsity of samples in the IHDP dataset, the joint optimization of $S$ and $Y$ leads to a slight improvement of performance. However, on the other three datasets, joint optimization $S$ and $Y$ has brought significant enhancements. Specifically, there is a $48.2\%$ improvement of $\mathbf{MSE_s}$ and $51.3\%$ of $\mathbf{MSE_y}$ in the Jobs dataset. As for Twins, the error of $S$ decreases $19.7\%$ and that of $Y$ decreases $33.3\%$. It yields even more significant improvements on the Simulation dataset, with a $76.6\%$ increase in accuracy for $S$ and $73.9\%$ for $Y$.

Joint CFR can be seen as a foundational version of our method, where the inputs of predicting $S$ and $Y$ only include the representation of $X$ and $T$. However, the overall objective function will consider the optimization directions of two counterfactual predictions and representation learning. For simplicity, we denote $+\hat{s}$ as an expanded version of Joint CFR, utilizing the predicted value of short-term outcome, i.e. $\hat{S}$, as an input to facilitate the prediction of long-term outcome $Y$. Considering that $Y$ already contains a part of information from $S$ in joint optimization, the improvement brought by using $S$ as an additional input is not very significant. Apart from the slight improvements in predicting $S$ on the IHDP dataset and $Y$ on the Simulation dataset, the performance gains in other prediction tasks are above $10\%$. Particularly noteworthy is the $25\%$ reduction in $\mathbf{MSE_y}$ on the Twins dataset and a $55.8\%$ reduction in $\mathbf{MSE_s}$ on the simulated dataset.

Actually, this method provides an initial solution of Pareto frontier exploration for our ultimate model, which is denoted as +$\hat{S}$+pareto. Such Pareto optimization is able to trade-off between the multiple conflicting objectives, including prediction of $S$ and $Y$, together with representation learning of $X$. In the IHDP dataset, the errors for predicting $S$ and $Y$ decrease by $22.7\%$ and $6.7\%$, respectively. Although there is limited enhancement in Twins dataset, the application of Pareto optimization leads to quite significant performance improvements in the other two datasets. To be specific, the declines in $\mathbf{MSE_s}$ and $\mathbf{MSE_y}$ reach $45.5\%$ and $55.6\%$ in Jobs dataset. As for the Simulation dataset, the complete version of our proposed method outperforms joint CFR + $\hat{S}$ with an improvement of $47.8\%$ in $\mathbf{MSE_s}$ and $77.3\%$ in $\mathbf{MSE_y}$.

According to the analysis above, the MSE error exhibits a monotonically decreasing pattern when analyzed row-wise. It validates the effectiveness of each newly introduced module, including the joint optimization among multiple objectives, the information of $\hat{S}$ as an input for prediction of $Y$, and Pareto optimization for further trade-off among conflicting objectives. Overall speaking, the first substantial improvement of estimation performance is attributed to the joint optimization between $S$ and $Y$, and another significant increase is after the application of Pareto optimization among the three objectives.

\subsubsection{Hyper-parameter}
Given several pairs of the initial weights for multiple tasks, the Pareto optimization algorithm will explore the corresponding Pareto frontiers that are locally continuous. We leave $20\%$ of the training data as a validation set to guide the selection of the final solution from them. The initial weights have great influence on the performance of counterfactual prediction, and so we conduct an experiment for further study. In our setting, $\alpha+\beta=1$ and we fix $\gamma$ as $0.001$ based on experience. In order to comprehensively explore the local Pareto frontiers, we adjust the initial values of $\alpha$ (i.e. $w_1$) as $0.1$, $0.3$, $0.5$, $0.7$, and $0.9$. Experimental results are reported in Table~\ref{tab:hyper}. Besides, we set $\rho=1$ in Eq. (\ref{eq:pareto}).

On the Twins dataset, the improvement of this Pareto optimization is not obvious since the best performance of each initial solution is very similar. However, this strategy on other three datasets plays a more important role, where different initial solutions lead to large difference in the performance of the trained model. In this case, dynamically adjusting the weight of multiple tasks and selecting the best solution shows superiority. As for the IHDP dataset, the final results achieved by our model are similar to the results with initial $\alpha=0.9$. It is easy to understand that our model mostly chooses the solution with $\alpha=0.9$ out of $10$ random seeds. However, it is likely that the optimal initial weights will also vary corresponding to different random seeds. For instance, the $\mathbf{MSE_y}$ on the Jobs dataset achieves a $19.9\%$ improvement, and the $\mathbf{MSE_s}$ on the Simulation dataset decreases by $65.7\%$.

In general, assigning higher weights to a specific objective can yield more accurate prediction of this single task but exacerbates the imbalance among multiple objectives. For example, there is a clear decreasing trend demonstrated in $\mathbf{MSE_s}$ but increasing trend in $\mathbf{MSE_y}$ on the IHDP dataset. However, being too focused on the balance among multiple tasks can also lead to local optimal solution. Like the results of Jobs, Twins, and Simulation datasets, the optimal solution may correspond to a more aggressive weight assignment. Therefore, utilizing the Pareto principle enables the exploration of more balanced or even superior results.

\section{Conclusion}
In this paper, we propose a novel end-to-end Pareto-Efficient framework with the purpose to determine the appropriate treatment value that would achieve Pareto optimality between short-term and long-term outcomes. Our framework consists of two key components: (1) Pareto-Optimal Estimation (POE) for predicting the potential outcomes in the short-term and long-term, and (2) Pareto-Optimal Policy Learning (POPL) for identifying the treatment value of Pareto optimality between multiple objectives. The backbones of both modules are designed and trained in a similar manner. In POE, we employ two regression losses and one representation loss to capture the prediction accuracy of the short-term and long-term potential outcomes. The POPL module leverages two regret losses, which act as guiding signals to determine the Pareto optimal treatment value. 

The key concern to solve is that within each module, all the losses conflict with each other during the optimization process. Therefore, we employ a continuous Pareto algorithm, which seeks to strike a balance between these different objectives. In this way, it can be ensured that no single objective will dominates the training procedure.

To evaluate the performance of our proposed method, we conduct extensive experiments on five datasets, including one real-life dataset, three semi-synthetic datasets, and one simulated dataset. Experimental results demonstrate the effectiveness of our approach. In counterfactual inference, our method significantly outperforms the sub-optimal baseline by notable improvements of $8.0\%\sim94.6\%$ in predicting $S$ and $25.1\%\sim94.3\%$ in predicting $Y$. As for policy learning, our method consistently resides on the Pareto frontier. We analyze the interpretability of these enhancements in treatment effect estimation, and demonstrate through ablation experiments that each module makes contributions, especially the joint optimization and Pareto optimization.





\ifCLASSOPTIONcaptionsoff
  \newpage
\fi



%

\bibliographystyle{IEEEtran}
\bibliography{example}

%

\begin{IEEEbiography}[{\includegraphics[width=1in,height=1.25in,clip,keepaspectratio]{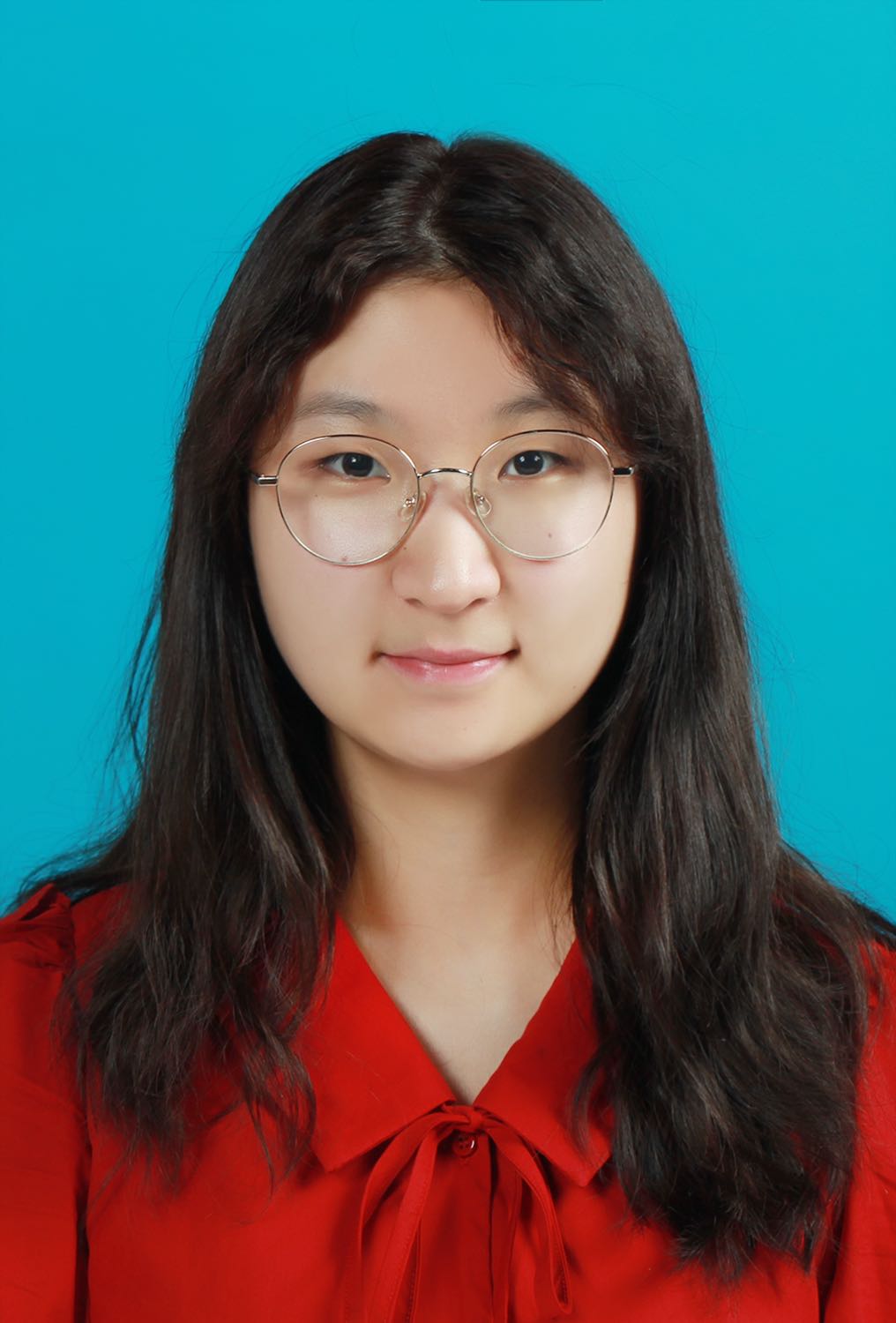}}]{Yingrong Wang} received the B.S. degree in 2021 from the College of Computer Science, Chongqing University. She is a third-year Ph.D. candidate in the College of Computer Science and Technology, Zhejiang University. Her current research interests include causal inference with short-term and long-term treatment effects, causal effect estimation of complex treatments, and policy learning.
\end{IEEEbiography}

\begin{IEEEbiography}[{\includegraphics[width=1in,height=1.25in,clip,keepaspectratio]{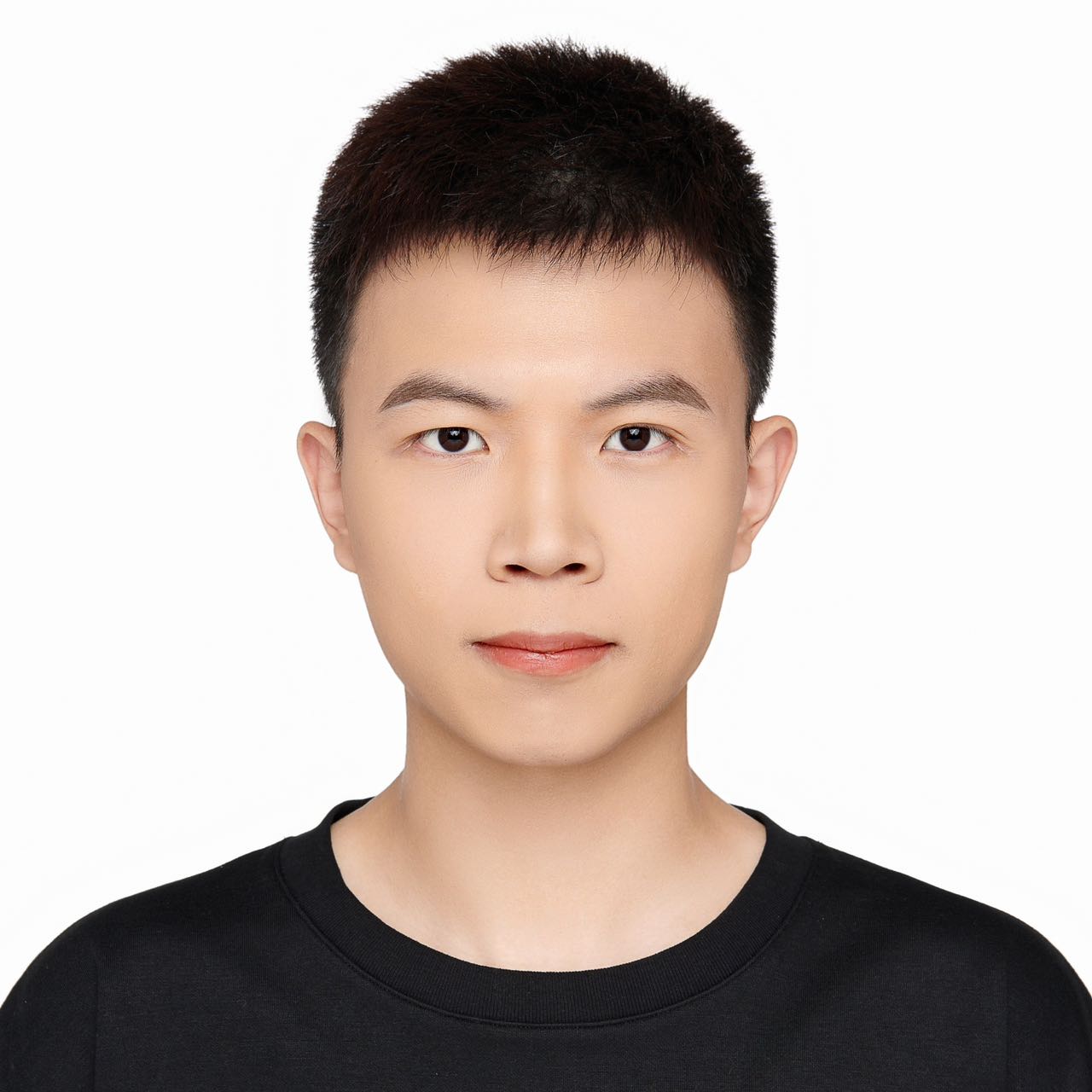}}]{Anpeng Wu} received the B.S. degree in 2020 from the College of Science, Zhejiang University of Technology. Currently, he is a fourth-year Ph.D. candidate in the Department of Computer Science and Technology of Zhejiang University. His main research interests include causal inference, representation learning and reinforcement learning.
\end{IEEEbiography}

\begin{IEEEbiography}[{\includegraphics[width=1in,height=1.25in,clip,keepaspectratio]{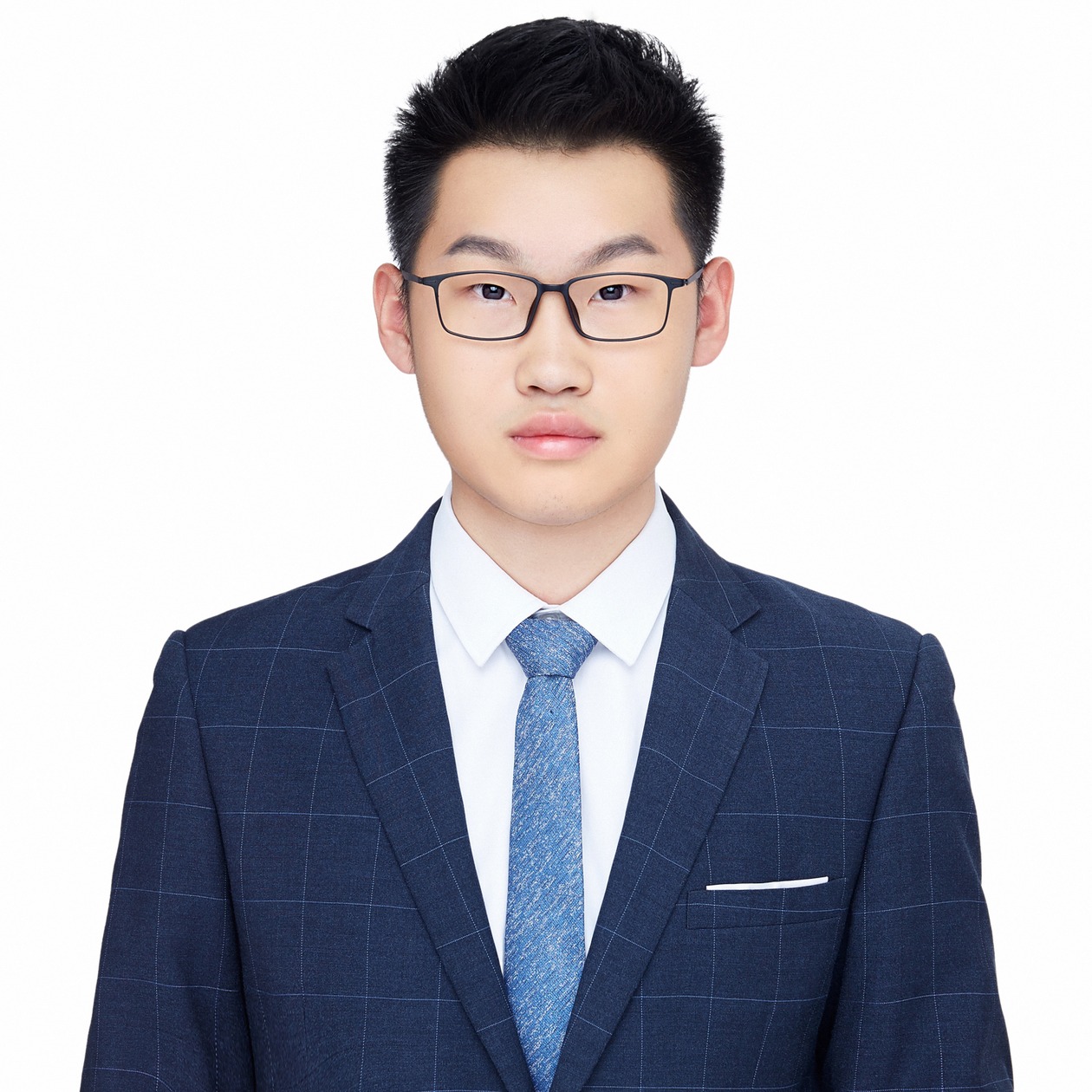}}]{Haoxuan Li} (Member, IEEE) received the B.S. degree from the School of Mathematics, Sichuan University. Currently, he is a third-year Ph.D. candidate in the Center for Data Science, Peking University. His main research interests include causal inference, recommendation system, and reinforcement learning.
\end{IEEEbiography}

\begin{IEEEbiography}[{\includegraphics[width=1in,height=1.25in,clip,keepaspectratio]{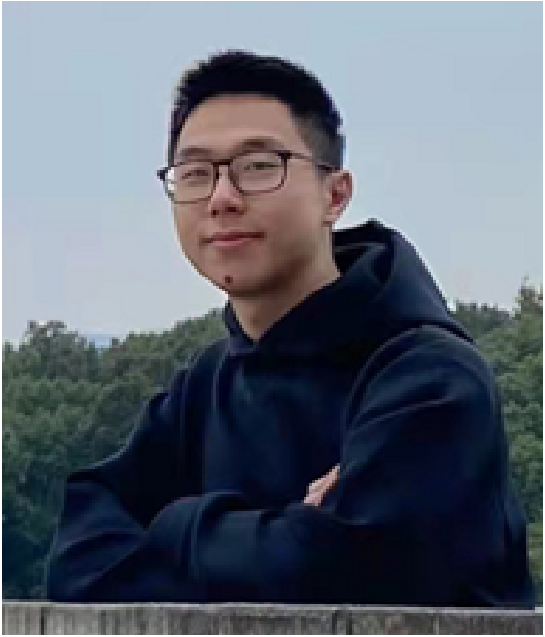}}]{Weiming Liu} is currently pursuing a Ph.D. degree at the College of Computer Science and Technology, Zhejiang University, Hangzhou, P.R. China. He received his graduate‘s degree in Electronic Science and Technology from Zhejiang University in 2021. His research interests include transfer learning with its applications on recommendation system.
\end{IEEEbiography}

\begin{IEEEbiography}[{\includegraphics[width=1in,height=1.25in,clip,keepaspectratio]{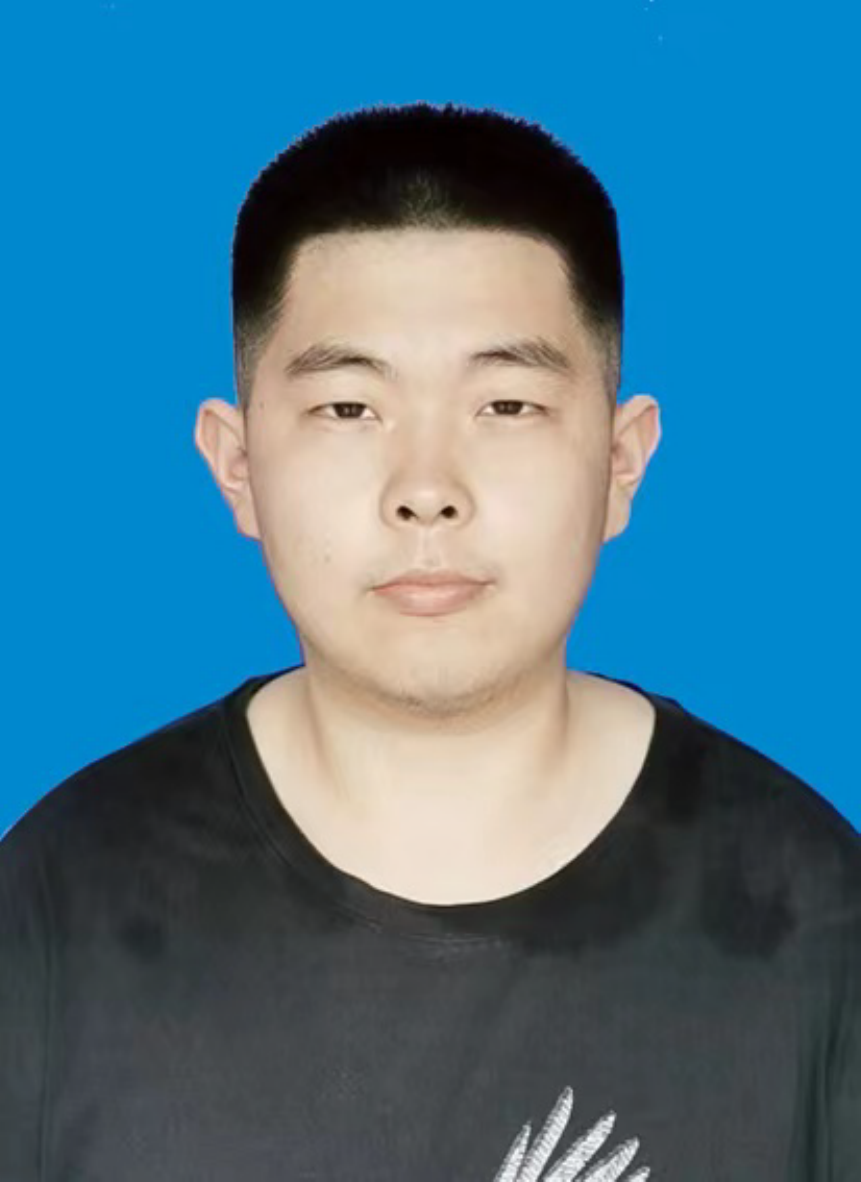}}]{Qiaowei Miao} received the B.S degree in 2021 from School of Cyber Security and Computer, Hebei University. He is currently pursuing the Ph.D. degree in the School of Software Technology, Zhejiang University. His main research interests include 3d gaussian splatting and text-guided diffusion.
\end{IEEEbiography}

\begin{IEEEbiography}[{\includegraphics[width=1in,height=1.25in,clip,keepaspectratio]{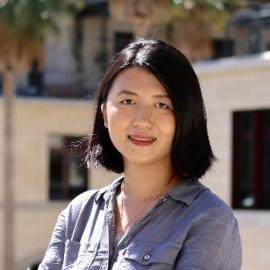}}]{Ruoxuan Xiong} received her Ph.D. in Management Science and Engineering from Stanford, advised by Markus Pelger. She was a postdoctoral fellow at the Stanford Graduate School of Business mentored by Susan Athey and Mohsen Bayati. Her research interests lie at the intersection of econometrics and operations management, focusing on causal inference, experimental design and factor modeling, with applications in finance and healthcare.
\end{IEEEbiography}

\begin{IEEEbiography}[{\includegraphics[width=1in,height=1.25in,clip,keepaspectratio]{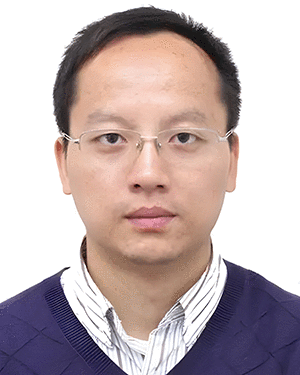}}]{Fei Wu} (Senior Member, IEEE) received the Ph.D. degree from Zhejiang University, Hangzhou, China. He was a Visiting Scholar with the Prof. B. Yu’s Group, University of California at Berkeley, Berkeley, from 2009 to 2010. He is currently a Full Professor with the College of Computer Science and Technology, Zhejiang University. His current research interests include multimedia retrieval, sparse representation, and machine learning.
\end{IEEEbiography}

\begin{IEEEbiography}[{\includegraphics[width=1in,height=1.25in,clip,keepaspectratio]{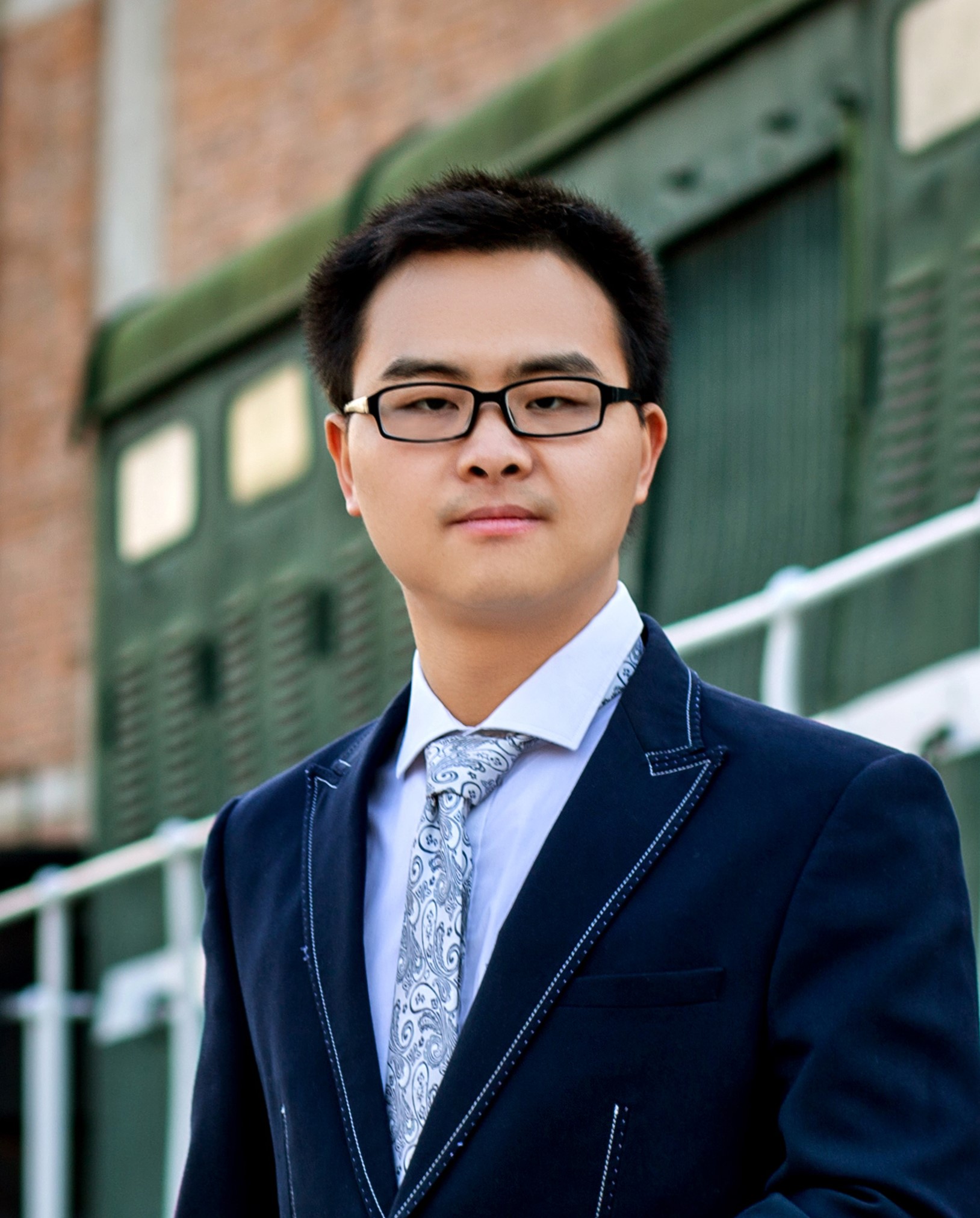}}]{Kun Kuang} received his Ph.D. degree from Tsinghua University in 2019. He is now an Associate Professor in the College of Computer Science and Technology, Zhejiang University. He was a visiting scholar with Prof. Susan Athey’s Group at Stanford University. His main research interests include Causal Inference, Artificial Intelligence, and Causally Regularized Machine Learning. He has published over 40 papers in major international journals and conferences, including SIGKDD, ICML, ACM MM, AAAI, IJCAI,
TKDE, TKDD, Engineering, and ICDM, etc.
\end{IEEEbiography}




\end{document}